\documentclass[lettersize,journal, dvipsnames]{IEEEtran}
\usepackage{amsmath,amsfonts}
\usepackage{algpseudocode}
\usepackage{algorithm}
\usepackage{array}
\usepackage[caption=false,font=normalsize,labelfont=sf,textfont=sf]{subfig}
\usepackage{textcomp}
\usepackage{stfloats}
\usepackage{url}
\usepackage{verbatim}
\usepackage{graphicx}
\usepackage{cite}
\usepackage{bm}
\usepackage{caption}
\usepackage{tikz}
\usepackage{subcaption}
\usepackage{arydshln}
\usepackage{booktabs}
\usepackage{siunitx}
\usepackage{newtxtext,newtxmath}
\newcommand{\dashedline}{%
  \noindent\leavevmode
  \leaders\hbox{{\color{gray}\rule[0.3ex]{0.2em}{0.1pt}}\hskip 0.3em}\hfill\kern0pt
}
\begin{document}

\title{Hierarchical Intention Tracking with Switching Trees for Real-Time Adaptation to Dynamic Human Intentions during Collaboration}
\author{Zhe Huang*, Ye-Ji Mun*, Fatemeh Cheraghi Pouria, and Katherine Driggs-Campbell
\thanks{* denotes equal contribution as the first author.}
\thanks{Z. Huang, Y. Mun, F. C. Pouria, and K. Driggs-Campbell are with the Department of  Electrical and Computer Engineering at the University of Illinois Urbana-Champaign. emails: \{zheh4, yejimun2, fatemeh5, krdc\}@illinois.edu}
}



\maketitle

\begin{abstract}
    During collaborative tasks, human behavior is guided by multiple levels of intentions that evolve over time, such as task sequence preferences and interaction strategies. To adapt to these changing preferences and promptly correct any inaccurate estimations, collaborative robots must accurately track these dynamic human intentions in real time. We propose a Hierarchical Intention Tracking (HIT) algorithm for collaborative robots to track dynamic and hierarchical human intentions effectively in real time. HIT represents human intentions as intention trees with arbitrary depth, and probabilistically tracks human intentions by Bayesian filtering, upward measurement propagation, and downward posterior propagation across all levels. We develop a HIT-based robotic system that dynamically switches between Interaction-Task and Verification-Task trees for a collaborative assembly task, allowing the robot to effectively coordinate human intentions at three levels: task-level (subtask goal locations), interaction-level (mode of engagement with the robot), and verification-level (confirming or correcting intention recognition). Our user study shows that our HIT-based collaborative robot system surpasses existing collaborative robot solutions by achieving a balance between efficiency, physical workload, and user comfort while ensuring safety and task completion. Post-experiment surveys further reveal that the HIT-based system enhances the user trust and minimizes interruptions to user's task flow through its effective understanding of human intentions across multiple levels. The video demonstrating our experiments is available at \url{https://youtu.be/Y5kg7QC41yw}.
\end{abstract}
\section{Introduction}
\label{sec:intro}
Robots require an effective understanding of human intentions to collaborate both safely and efficiently with humans. During long-term tasks, human intentions continuously evolve along with task progress. When handling a complex task, humans typically break down the task into milestones and sub-tasks at varying levels of granularity, leading to a hierarchical structure of human intentions. During collaboration, humans often  maintain multiple intentions with different semantics simultaneously. For instance, they may prefer specific subtask sequences or modes of interaction with the robot (e.g. hand guidance or autonomous control), and these intentions can change over time.

Despite the inherent complexity and dynamics of human intention during collaboration, prior works typically model human intention as a single random variable that does not evolve over time~\cite{cacace2018shared, mascaro2023intention}. This simplified definition potentially causes premature convergence of intention estimation before task transitions. Another alternative approach is single-shot human intention classification~\cite{driggs2015identifying}, which infers intentions based on a fixed observation window. However, such methods are constrained by incomplete history context, resulting in limited inference accuracy. Moreover, existing works are mainly focused on modeling human intentions at a single level~\cite{geravand2013human, nicolis2018human}, neglecting the hierarchical structure of human intentions and the mutual influence between different levels of human intentions. Estimating intentions as a single level (i.e., without a hierarchy) degrades the temporal resolution and semantic granularity of intentions. Such simplification compromises downstream collaborative robot execution pipelines, which rely on temporally and semantically structured intention signals to generate context-appropriate actions.

\begin{figure}
    \centering
    \includegraphics[width=1\linewidth]{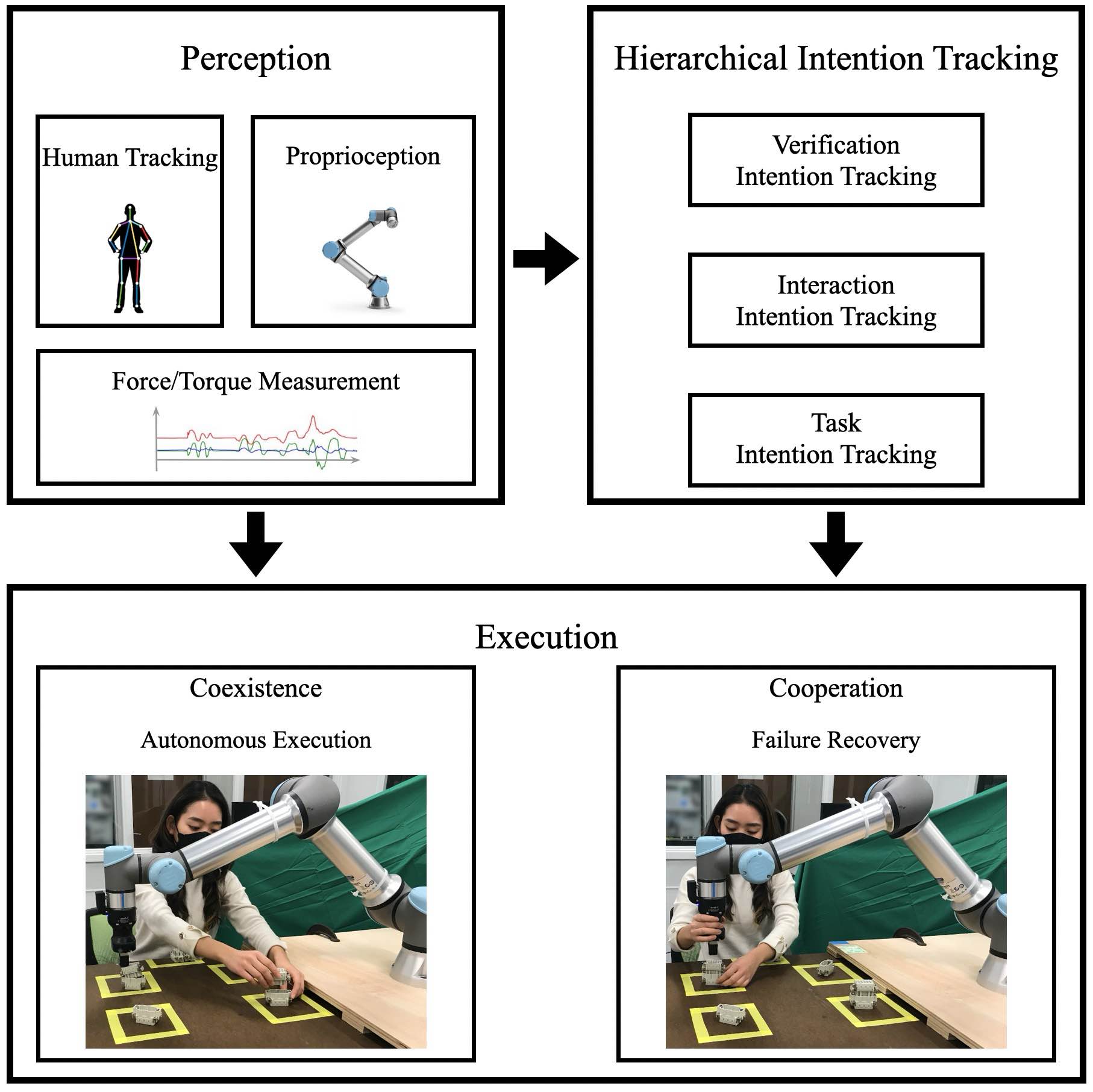}
    \caption{\textbf{Overview of Hierarchical Intention Tracking (HIT) based collaborative robot system.} The robot continuously tracks evolving human intentions at task, interaction, and verification levels by integrating human body tracking, proprioception (i.e. awareness of its own joint positions and motion), and force/torque measurements. The robot accordingly aligns its behavior with the human collaborator by feeding the tracked human intentions to downstream execution: autonomously performing assembly with human-aware collision avoidance in Coexistence Mode and offering compliant manual guidance for failure recovery in Cooperation Mode.}
    \label{fig:hit-system-overview}
    \vspace{-0.5cm}
\end{figure}

In this work, we introduce Hierarchical Intention Tracking (HIT), an algorithm that equips the robot with the capability to understand dynamic, multi-level human intentions during long-term, complex collaborative tasks. We formulate a probabilistic graphical model, where human intentions evolve as a Markov process. Leveraging Bayesian filtering, our approach tracks human intention in real time, conditioned on the history of observable human and robot states. To capture hierarchical complexity, we propose modeling human intentions through intention trees with arbitrary depth, employing upward measurement propagation and downward posterior propagation to integrate information across different levels effectively. This framework can be adapted to different deployments by adjusting the number of intention trees, their depth, and the number of semantic intentions considered.

We validate the effectiveness of HIT in a free-form collaborative assembly task, where a human and a robot work together to assemble multiple male and female parts within designated regions as shown in Figure~\ref{fig:hit-system-overview}. The human aligns the male and female parts, and the robot pushes two parts together to finish the assembly. By tracking the human intention at task level, the robot plans the sequence of assembly tasks according to the part alignment order performed by the human. At the interaction level, the robot dynamically switches between executing autonomous assembly actions (\textit{Coexistence}) or reaching towards the human to offer manual guidance mode (\textit{Cooperation}). This interaction mode switching is based on tracking the human intention at interaction level, whether the human is working independently on part alignment or seeking assistance from the robot for failure recovery. At verification level, the robot continuously validates human interaction-level intentions and reverts to autonomous mode (\textit{Coexistence}) if the human is incorrectly inferred to seek manual guidance (\textit{Cooperation}), thereby ensuring safety. 

We build a HIT-based human-robot collaboration (HRC) system featuring two intention trees: Interaction-Task (IT) tree and Verification-Task (VT) tree. We implement Coexistence and Cooperation Modules to support different interaction modes. Importantly, our HIT framework is general and configurable, allowing deployment in arbitrary collaboration scenarios with varying number of trees, tree depths, and semantic intention levels. Our comprehensive user study shows that our HIT-based HRC system significantly outperforms existing baselines and
achieves seamless, robust, and user-friendly collaboration. 

Our key contributions are summarized as follows:
\begin{itemize}
    \item We propose a \textbf{Hierarchical Intention Tracking (HIT)} algorithm that estimates dynamic, multi-level human intentions using structured intention trees in real time.
    
    \item We implement and evaluate the \textbf{HIT-based collaboration robot system with Interaction-Task and Verification-Task trees (HIT-ITVT)} in a free-form assembly task, demonstrating significant improvements over baselines in task coordination, robustness in failure recovery, verifying tracked intentions, and user experience.

    \item Our user study shows that \textbf{HIT-ITVT} outperforms baselines in terms of satisfaction, trust, adaptability, intention alignment, and human fatigue reduction, and underscores the benefits of hierarchical intention tracking and self-correction in HRC.
\end{itemize}

The remaining part of this work is organized as follows. In Section~\ref{sec:related}, we review the related works on human intention modeling and HRC. In Section~\ref{sec:hit}, we formulate the intention tracking problem in generic HRC and introduce the HIT algorithm. In Section~\ref{sec:hit++}, we introduce our implementation of HIT-based HRC system, HIT-ITVT, for a collaborative assembly task. Section~\ref{sec:implementation} provides the implementation details of our collaborative assembly use case. In Section~\ref{sec:userstudy}, we present our result of the user study evaluating the performance of our HIT-ITVT in the collaborative assembly task and illustrate the effectiveness of HIT in HRC. Finally, we conclude with a summary of our contributions on HIT and discussion on future directions of this work in Section~\ref{sec:conclusion}.

\section{Related Works}
\label{sec:related}
\subsection{Intention Inference}
Intention recognition has been widely studied as a means of improving robot assistance and coordination in HRC. The study of intention recognition has been applied across a variety of domains, including autonomous navigation~\cite{rasouli2019pie, liu2023intention}, assistive robotics~\cite{patel2022proactive, chang2023specifying}, and collaborative manipulations~\cite{li2013human,alharthi2024towards}. A broad range of sensing technologies have been explored to capture human intention—particularly non-verbal cues—through modalities such as vision-based tracking, gaze estimation, and force sensing. Vision-based methods are commonly used to infer intentions by analyzing human body movements, gestures, and actions and modeling intention as a function of observable behaviors~\cite{zhuang2022goferbot, breazeal2005effects, luo2019human}. Recent advancements further incorporate human-object interaction to provide contextual cues about task goals, since interactions with different objects can signal distinct intentions~\cite{adebayo2024qub, mascaro2023hoiabot}. Gaze behavior has also been proven to be a strong predictor of intention, used both to anticipate human needs~\cite{chang2023specifying} and as an early indicator of forthcoming actions~\cite{huang2015using}. In physical human-robot interaction tasks such as teleoperation or co-manipulation,  contact force applied by the human is often translated into intention estimates~\cite{cacace2018shared, li2015continuous, alharthi2024towards}. These studies collectively highlight the importance of multimodal sensing to effectively capture the subtleties of human communication and behavior. Building on this insight, our work integrates gesture and contact force to infer human intention in a contextualized manner.

Intention inference methods can be categorized into several major approaches: probabilistic models, learning-based, and context-aware methods. Probabilistic models—particularly Bayesian filtering—are widely adopted for their real-time adaptability and capacity to model uncertainty and temporal dynamics~\cite{wang2013probabilistic, liu2018goal, jain2018recursive}. These methods update beliefs about human intentions as new observations become available. In parallel, neural network-based methods, such as Long Short-Term Memory (LSTM)~\cite{huang2021long, zhang2022prediction} and Radial Basis Function (RBF) networks~\cite{liu2019intention}, have been applied for motion and velocity prediction in human-robot interaction. In addition, context-aware methods incorporate scene understanding and human goals to identify relevant key features~\cite{cai2024hierarchical, adebayo2024qub}. Many of these methods often model human behaviors under Gaussian assumptions to enable tractable inferences~\cite{wang2013probabilistic, zhang2022prediction}. Building on this body of work, we employ Bayesian inference combined with Gaussian behavior modeling to enable real-time multi-intention tracking using contextual information.

Due to the inherent complexity and evolving nature of human behavior, hierarchical intention models have gained attention for their ability to represent multi-level structures and adapt to changes over time~\cite{cai2024hierarchical, cheng2020towards}. Unlike flat models, hierarchical approaches allow continuous updates across abstraction levels, improving robustness in dynamic tasks~\cite{cacace2018interactive}. For instance, AND-OR graphical models~\cite{hawkins2014anticipating} and Bayesian goal inference~\cite{liu2018goal, cheng2020towards} have demonstrated improved performance in noisy or ambiguous environments. These hierarchical probabilistic models also enable real-time management of multiple human intentions and task preferences by continuously updating multi-level estimates as the task progresses~\cite{huang2023hierarchical}. 
However, as hierarchical models may still generate incorrect estimates in the presence of novel or noisy behavior, control architectures must incorporate mechanisms to detect and recover from such mispredictions~\cite{chen2022insights}. Some approaches learn a human’s potentially imperfect internal dynamics through inverse reinforcement learning, enabling the robot to adapt its assistance accordingly~\cite{reddy2018you}. In this work, we address this challenge by proposing a HIT framework equipped with a verification mechanism to ensure robust performance in dynamic, human-centered environments.

\subsection{Human-Robot Collaboration}
HRC can be categorized into three levels, each reflecting an increasing degree of interactivity and sophistication in how robots engage with humans. The most fundamental level is safety-based interaction, where their primary function is to ensure physical safety by treating humans as dynamic obstacles and avoiding collision~\cite{khatib1986real}. This safety functionality forms the foundation of all collaborative systems operating in shared workspaces. The second level involves task-aware coordination, where the robot interprets human motion and recognizes the task being performed. This level includes understanding human actions using techniques ranging from Kalman filters~\cite{kohler1997using, bruce2004better}, to graphical models~\cite{wang2013probabilistic, liu2023intention} and deep learning models that process sensory inputs~\cite{zhang2022prediction, liu2019intention}. Robots at the second level adjust their behavior to remain aligned with human task execution. The highest level is intention-aware collaboration, where the robot goes beyond task recognition to infer high-level human intentions and adapts its behavior in real time to reflect human preferences~\cite{adebayo2024qub, hernandez2024bayesian}. In this work, we advance intention-aware collaboration by developing a system that jointly considers human and robot states to track human intentions, enable failure recovery, and correct misdetections, resulting in more responsive and human-aligned robot behavior. 

HRC has been applied across a wide range of industrial tasks, including tool handover~\cite{perez2015fast, cheng2020towards}, heavy object lifting~\cite{grahn2016potential, kim2017anticipatory}, surface polishing~\cite{wilbert2012robot}, welding~\cite{shi2012levels}, and assembly~\cite{tsarouchi2017human, heydaryan2018safety}. These applications typically fall under one of three major automation modes: Safeguard, Coexistence, and Cooperation~\cite{de2012integrated, villani2018survey}, which correspond closely to all levels of HRC discussed earlier. In Safeguard Mode, the human and the robot are not allowed to move at the same time within shared work space; robot motion is suspended when the human is detected and resumes after the area is clear of humans~\cite{svarny2019safe, papanastasiou2019towards}. Coexistence Mode allows human and robot to work concurrently in close proximity, executing their respective tasks independently, with the robot actively avoiding interference or contact with its human partner~\cite{flacco2012depth,lasota2014toward, chen2018collision}. Cooperation Mode, by contrast, involves physical coordination, often relying on manual guidance or shared control from the human to accomplish tasks beyond the robot’s autonomous capabilities~\cite{geravand2013human,cacace2018shared, nicolis2018human}. In alignment with the task-aware coordination level of HRC, our work is focused on enabling effective collaboration in free-form assembly environments by combining elements of both Coexistence and Cooperation Mode for concurrent operation and failure recovery, respectively.

\section{Hierarchical Intention Tracking}
\label{sec:hit}

We propose the HIT algorithm to track changing human intentions at multiple levels during HRC. In this section, we first formulate the problem of human intention tracking in the HRC setting, then derive intention tracking in a single intention layer setting, and finally extend to HIT with an arbitrary number of intention levels.

\begin{figure}[t!]
    \centering
    \includegraphics[width=0.8\linewidth]{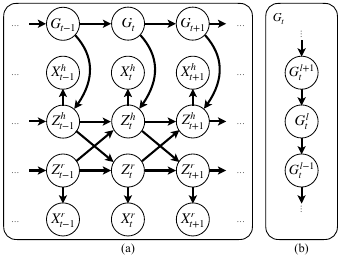}
    \caption{\textbf{Illustration of probabilistic graphical models.} (a) A probabilistic graphical model of intention-evolving HRC. We denote observed human and robot states as $X^h_t$ and $X^r_t$ respectively, and their corresponding latent states are represented by $Z^h_t$ and $Z^r_t$. The human intention is denoted by~$G_t$. (b) A probabilistic graphical model of hierarchical intentions. This model depicts hierarchy of intentions, where $G_t^l$ represents the intention at level-$l$. The bottom level intention $G_t^{1}$ is a parent of $Z^h_t$ in (a).}
    \label{fig:hit_pgm}
    \vspace{-0.3cm}
\end{figure}

\subsection{Problem Formulation}
Consider a collaborative task with multiple subtask goals, which involves a team of one human and one robot working together. The human leads the team by determining the desired sequence of goals, while the robot continuously tracks the human's intended goal to provide timely and appropriate assistance. We define the human intention at time $t$, denoted by $G_t$, as a discrete random variable representing the human's desired goal. The human-robot team dynamics are formulated as a probabilistic graphical model shown in Figure~\ref{fig:hit_pgm}(a). We assume the human intention $G_t$ follows a Markov process, as the human leads the team and determines task goals independent of the robot actions. The latent human state $Z^h_{t}$ is influenced by the human intention $G_t$, while the latent human state $Z^h_{t}$ and the latent robot state $Z^r_{t}$ are also conditioned on both latent states from previous time step $Z^{h,r}_{t-1}$, reflecting the interaction dynamics during collaboration. Ideally, seamless HRC is achieved when the robot behavior consistently aligns with the ground-truth human intention. To enable this alignment, we aim to infer and track the human intention $G_t$ in real time based on the observable state history of both the human and the robot $X^{h,r}_{1:t}$.

When tasks become more complicated, a hierarchical structure often emerges in the human goals, arising from the task decomposition of high-level milestones into low-level to-do items. Such collaborative tasks also require reasoning over multiple types of human intentions—not only the task sequence, but also interaction modes and considerations for personal safety. In these scenarios, the human intention $G_t$ is structured hierarchically as presented in Figure~\ref{fig:hit_pgm}(b). The intention at layer $l$ and time $t$ is represented by $G_t^l$ and is tracked based on the observable state history $X^{h,r}_{1:t}$.

\subsection{Single-layer Intention Tracking for HRC}

\begin{figure}[b]
    \centering
    \vspace{-0.2cm}
    \includegraphics[width=\linewidth]{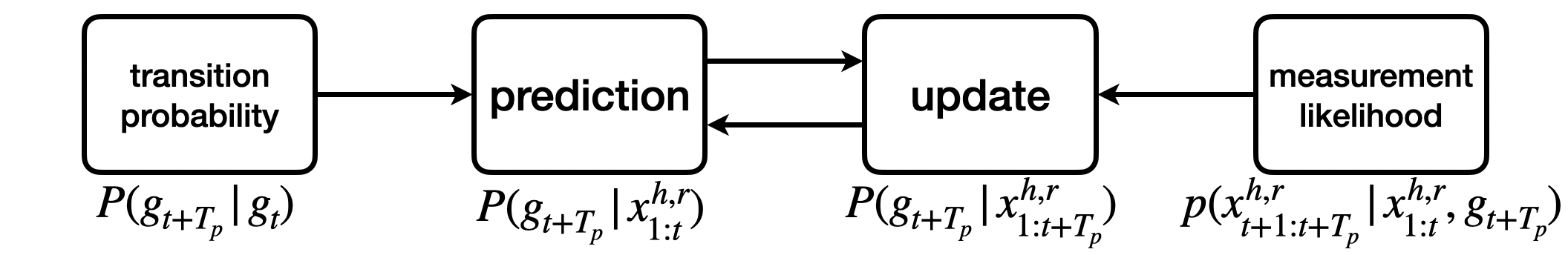}
    \caption{\textbf{Single-layer Intention Tracking.} Human intention $g_t$ is estimated using Bayesian filtering based on the observable human and robot state history $x_{1:t}^{h, r}$. The parameter $T_p$ defines the number of time steps considered for each prediction and update iteration for intention tracking.}
    \label{fig:hit_single_layer}
\end{figure}

\begin{figure*}[hb!]
    \centering
    \vspace{-0.7cm}
    \includegraphics[width=0.8\linewidth]{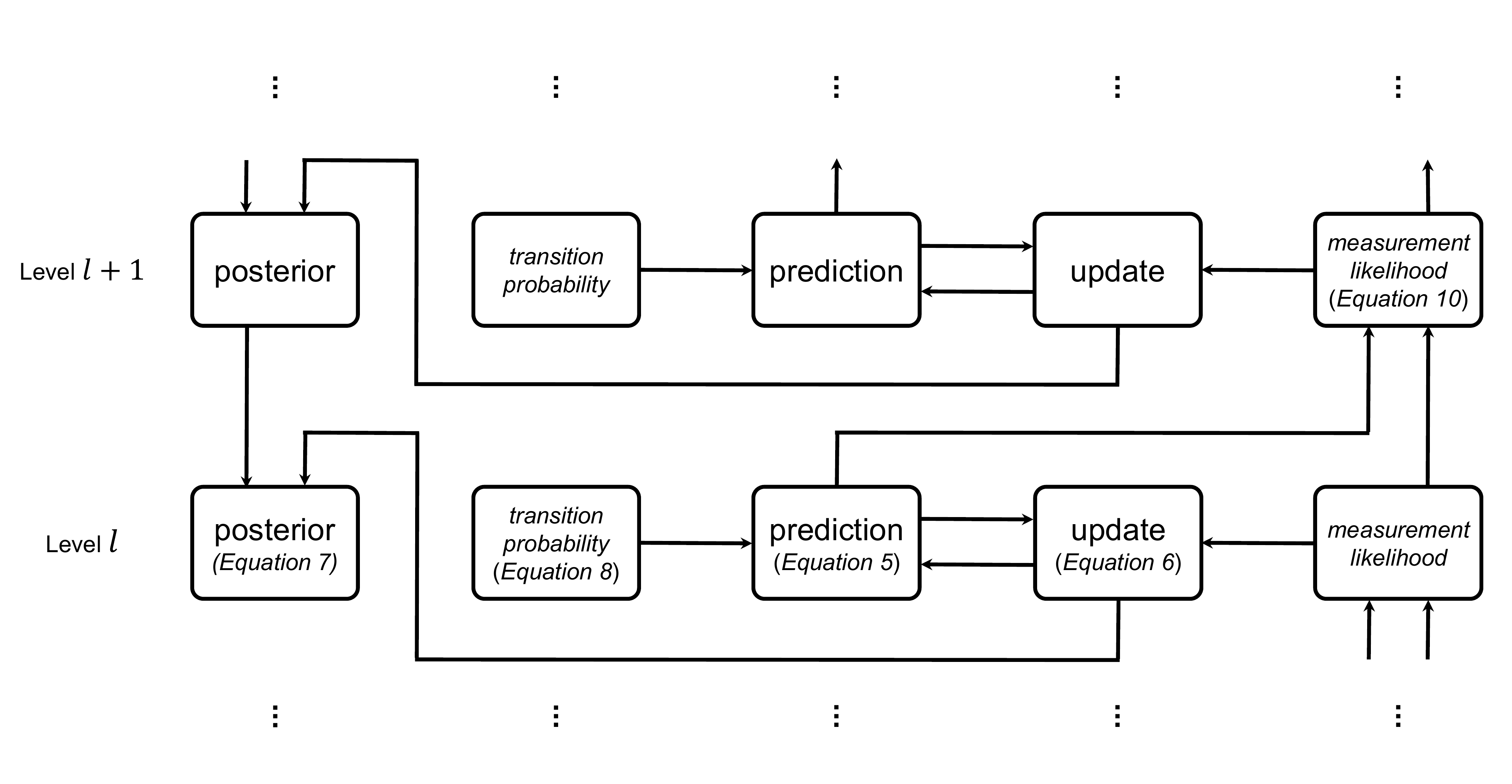}
    \caption{\textbf{Illustration of Hierarchical Intention Tracking between adjacent intention levels.}}
    \label{fig:hierarchical_intention_tracking_algorithm_illustration}
\end{figure*}

We first introduce intention tracking when the human intention $G_t$ has a single layer represented by a single discrete random variable. As shown in Figure~\ref{fig:hit_single_layer}, we track human intentions in every $T_p$ time step by Bayesian filtering to obtain the posterior over $G_t$ conditioned on observable state history~$X_{1:t}^{h,r}$. The prediction and update steps of this filtering process are presented in Equation~\ref{eq:hit_single_layer_prediction} and~\ref{eq:hit_single_layer_update} respectively.

\begin{equation}\label{eq:hit_single_layer_prediction}
P(g_{t+T_p} | x_{1:t}^{h,r}) = \sum_{g_{t}} P(g_{t+T_p} | g_{t}) P(g_{t} | x_{1:t}^{h,r})
\end{equation}
\begin{equation}\label{eq:hit_single_layer_update}
P(g_{t+T_p} | x_{1:t+T_p}^{h,r})\! \propto\!p(x_{t+1:t+T_p}^{h,r} | x_{1:t}^{h,r}, g_{t+T_p}) P(g_{t+T_p} | x_{1:t}^{h,r})
\end{equation}

We define the transition model $P(g_{t+T_p} | g_{t})$ in Equation~\ref{eq:hit_single_layer_prediction} as a time-invariant matrix
\begin{equation}\label{eq:hit_single_layer_transition_model}
P(g_{t+T_p}|g_{t}) = \begin{cases}
\kappa & \text{if $g_{t+T_p} = g_{t}$};\\
(1-\kappa)/(m-1), & \text{otherwise}.
\end{cases}
\end{equation}
where $\kappa$ is the probability of the human staying in the same intention after $T_p$ time steps, and $m$ is the cardinality of the sample space of $G_t$. A larger $\kappa$ implies that the human is more likely to keep the current intention. The intention duration $T_p$ may vary depending on the type of interaction. The probability of the human shifting to another intention is assumed equivalent as indicated in Equation~\ref{eq:hit_single_layer_transition_model}.

The measurement model $p(x_{t+1:t+T_p}^{h,r} | x_{1:t}^{h,r}, g_{t+T_p})$ in Equation~\ref{eq:hit_single_layer_update} can be represented by an intention-aware time-series human behavior model as in Equation~\ref{eq:hit_single_layer_measurement_model} with two assumptions: (1) the human intentions during the tracking iteration period $g_{t+1:t+T_p}$ are consistent; (2) the robot follows a deterministic policy conditioned on the human intention. 
\begin{equation}\label{eq:hit_single_layer_measurement_model}
p(x_{t+1:t+T_p}^{h,r} | x_{1:t}^{h,r}, g_{t+T_p})\!\!=\!\!\prod_{\tau=1}^{T_p} \!\!p(x_{t+\tau}^{h} | x_{1:t+\tau-1}^{h,r}, x_{t+\tau}^r, g_{t+\tau})
\end{equation}

\subsection{Hierarchical Intention Tracking for HRC} \label{subsec:hit}

\begin{figure}[b]
    \centering
    \vspace{-0.4cm}
    \includegraphics[width=0.6\linewidth]{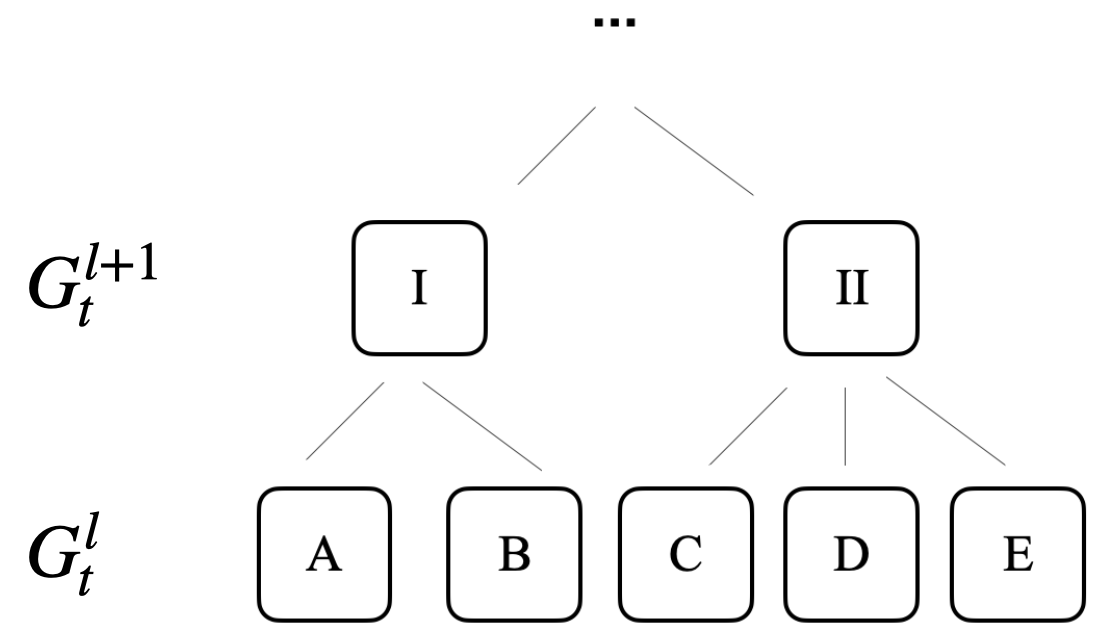}
    \caption{\textbf{An example of intention hierarchy which is represented by an intention tree.} Each intention has one and only one parent intention.}
    \label{fig:hit_intention_tree_example}
\end{figure}

We extend single-layer intention tracking to hierarchical intention tracking by considering adjacent layers ${G}_t^l$ and ${G}_t^{l+1}$ in the intention hierarchy as illustrated in Figure~\ref{fig:hierarchical_intention_tracking_algorithm_illustration}. We formulate the realization of the intention hierarchy as a tree structure, where each intention at level $l$ has one and only one parent intention at level $l\!+\!1$, as illustrated in Figure~\ref{fig:hit_intention_tree_example}.

The prediction and update steps for the single-layer intention tracking in Equation~\ref{eq:hit_single_layer_prediction} and Equation~\ref{eq:hit_single_layer_update} are extended to the level-$l$ intention with an additional condition on their parent intention $\textrm{Pa}(g_t^l)$ at level $l\!+\!1$ as follows:

\begin{equation}\label{eq:hit_prediction_conditioned_on_parent_intention}
    \begin{aligned}
        & P(g^l_{t+T_p} | x_{1:t}^{h,r}, \textrm{Pa}(g^l_{t+T_p})) = \\ 
        & \sum_{g^l_{t} \in \textrm{Ch}(\textrm{Pa}(g^l_{t+T_p}))} \mkern-40mu P(g^l_{t+T_p} | g^l_{t}, \textrm{Pa}(g^l_{t})) P(g^l_{t} | x_{1:t}^{h,r}, \textrm{Pa}(g^l_{t})) \\
    \end{aligned}
\end{equation}

\vspace{-0.1cm}

\begin{equation}\label{eq:hit_update_conditioned_on_parent_intention}
    \begin{aligned}
    & P(g^l_{t+T_p} | x_{1:t+T_p}^{h,r}, \textrm{Pa}(g^l_{t+T_p})) \propto \\
    & p(x_{t+1:t+T_p}^{h,r} | x_{1:t}^{h,r}, g^l_{t+T_p},\!\textrm{Pa}(g^l_{t+T_p})) P(g^l_{t+T_p} | x_{1:t}^{h,r},\!\textrm{Pa}(g^l_{t+T_p})) \\
    \end{aligned}
\end{equation}
where $\textrm{Ch}(\textrm{Pa}(g^l_{t+T_p}))$ denotes the set of child intentions of the parent intention of $g^l_{t+T_p}$, i.e., the sibling intentions of $g^l_{t+T_p}$ and $g^l_{t+T_p}$ itself which share the same parent intention.

The conditional posterior $P(g^l_{t+T_p} | x_{1:t+T_p}^{h,r}, \textrm{Pa}(g^l_{t+T_p}))$ from Equation~\ref{eq:hit_update_conditioned_on_parent_intention} handles the dynamics between the sibling intentions of $g^l_{t+T_p}$ and the common parent intention $\textrm{Pa}(g^l_{t+T_p})$ at level-$l\!+\!1$ intention. To obtain the posterior $P(g^l_{t+T_p} | x_{1:t+T_p}^{h,r})$ for HIT, we propose \textbf{downward posterior propagation}, which propagates the conditional posterior from the top level to the bottom level:

\begin{equation}\label{eq:hit_downward_posterior_propagation}
    \begin{aligned}
    & P(g^l_{t+T_p} | x_{1:t+T_p}^{h,r}) \\
    & = P(g^l_{t+T_p} | x_{1:t+T_p}^{h,r}, \textrm{Pa}(g^l_{t+T_p})) P(\textrm{Pa}(g^l_{t+T_p}) | x_{1:t+T_p}^{h,r})\\
    \end{aligned}
\end{equation}

At the top level $L$, the intention has no parent, so the conditional posterior is equivalent to the posterior $P(g^L_{t+T_p} | x_{1:t+T_p}^{h,r})$. Consider $g^l_{t+T_p}\!\!=\!\!\textrm{C}$ in Figure~\ref{fig:hit_intention_tree_example} as an example. The level-$l$ prediction and update steps conditioned on the parent intention in Equation~\ref{eq:hit_prediction_conditioned_on_parent_intention} and Equation~\ref{eq:hit_update_conditioned_on_parent_intention} yield the conditional posterior $P(g^l_{t+T_p}\!\!=\!\!\textrm{C} | x_{1:t+T_p}^{h,r}, \textrm{Pa}(g^l_{t+T_p})\!\!=\!\!\textrm{II})$. This conditional posterior captures the distribution over $\{\textrm{C}, \textrm{D},\textrm{E}\}$ under the condition that the parent intention is $\textrm{II}$. The downward posterior propagation step in Equation~\ref{eq:hit_downward_posterior_propagation} incorporates the parent posterior over $\{\textrm{I},\textrm{II}\}$ at level $l+1$ to obtain the posterior over the level-$l$ intentions $\{\textrm{A}, \textrm{B}, \textrm{C}, \textrm{D}, \textrm{E}\}$.

We define the transition model for level-$l$ intention conditioned on the parent intention in Equation~\ref{eq:hit_prediction_conditioned_on_parent_intention} as follows:

\begin{equation}\label{eq:hit_transition_model_conditioned_on_parent_intention}
    P(g^l_{t+T_p}|g^l_{t}, \textrm{Pa}(g^l_t)) = \begin{cases}
    \kappa^l, & \text{if $g_{t+T_p} = g_{t}$};\\
    \frac{1-\kappa^l}{m_{\textrm{Pa}(g^l_t)}-1}, & \text{elif $\textrm{Pa}(g^l_{t+T_p}) = \textrm{Pa}(g^l_{t})$};\\
    0 & \text{otherwise}. \\
    \end{cases}
\end{equation}
where the number of sibling intentions under the same parent is denoted as $m_{\textrm{Pa}(g^l_t)}$, and $\kappa^l$ is the probability of staying in level-$l$ intention.

The parent-conditioned measurement likelihood for a level-$l$ intention in Equation~\ref{eq:hit_update_conditioned_on_parent_intention} can be simplified by omitting the parent intention. This simplification is justified by the single-parent tree structure of the intention hierarchy, where the child intention $g^l_{t+T_p}$ encodes the information of its parent intention. As a result, the parent intention does not contribute additional information for the measurement likelihood:
\begin{equation}\label{eq:hit_measurement_model_conditioned_on_parent_intention_simplified}
    p(x_{t+1:t+T_p}^{h,r} | x_{1:t}^{h,r},\!g^l_{t+T_p},\!\textrm{Pa}(\!g^l_{t+T_p}\!))\!=\! p(x_{t+1:t+T_p}^{h,r} | x_{1:t}^{h,r},\!g^l_{t+T_p})
\end{equation}

To obtain the level-$l\!+\!1$ measurement likelihood, we propose \textbf{upward measurement propagation}, which aggregates the level-$l$ measurement likelihood based on the level-$l$ conditional prediction $P(g_{t+T_p}^l|x_{1:t}^{h,r}, \textrm{Pa}(g^l_{t+T_p}))$: 

\begin{equation}\label{eq:hit_upward_measurement_propagation_new}
    \begin{aligned}
        & p(x_{t+1:t+T_p}^{h,r} | x_{1:t}^{h,r},g^{l+1}_{t+T_p}) = \\ 
        &\sum_{g^l_{t+T_p} \in \,\textrm{Ch}(g^{l+1}_{t+T_p})}\!\!\!\!\!\!\!\!\!\!\!p(x_{t+1:t+T_p}^{h,r} | x_{1:t}^{h,r},g^{l}_{t+T_p}) P(g^{l}_{t+T_p} | x^{h,r}_{1:t}, g^{l+1}_{t+T_p})
    \end{aligned}
\end{equation}
where $\textrm{Ch}(g^{l+1}_{t+T_p})$ denotes the set of child intentions of the given parent intention $g^{l+1}_{t+T_p}$.

Note the bottom-level measurement model with the most concrete intention definition $p(x_{t+1:t+T_p}^{h,r} | x_{1:t}^{h,r},g^{1}_{t+T_p})$ is represented by the human behavior model in Equation~\ref{eq:hit_single_layer_measurement_model}. The proposed algorithm is summarized in Algorithm~\ref{algo:hit}.

\begin{algorithm*}[ht!]
    \caption{Hierarchical Intention Tracking}\label{algo:hit}
    \begin{algorithmic}[1]
    % \State Conditional Prior $P(g^l_0 | \text{Pa}(g^l_0)) \!\!\gets \!\!1/m_{\text{Pa}(g^l_0)}, \,\,l = 1, \ldots, L$,  
    \State $P(g^l_0 | \text{Pa}(g^l_0)) \gets 1/m_{\text{Pa}(g^l_0)}, \,\,l = 1, \ldots, L$; $t \gets 0$;
    \For {each $T_p$ time steps}
        \State Receive observable human and robot states $x^{h,r}_{t+1:t+T_p}$;
        \For {$l = 1$ to $L$}
            \If {$l = 1$} 
                \State Compute the bottom level measurement likelihood $p(x_{t+1:t+T_p}^{h,r} | x_{1:t}^{h,r},\!g^1_{t+T_p})$  in Equation~\ref{eq:hit_single_layer_measurement_model};
            \Else
                \State Perform upward measurement propagation $p(x_{t+1:t+T_p}^{h,r} | x_{1:t}^{h,r},\!g^l_{t+T_p})$ from level $l\!-\!1$ in Equation~\ref{eq:hit_upward_measurement_propagation_new};
            \EndIf

            \State Compute level-$l$ conditional prediction $P(g^l_{t+T_p} | x_{1:t}^{h,r}, \textrm{Pa}(g^l_{t+T_p}))$ in Equation~\ref{eq:hit_prediction_conditioned_on_parent_intention};
            
            \State Compute level-$l$ conditional posterior $P(g^l_{t+T_p} | x_{1:t+T_p}^{h,r}, \textrm{Pa}(g^l_{t+T_p}))$ in Equation~\ref{eq:hit_update_conditioned_on_parent_intention};
            
        \EndFor
        \For {$l = L$ to $1$}
            \If {$l = L$} 
                \State Set the top level posterior $P(g^L_{t+T_p} | x_{1:t+T_p}^{h,r})$ as the top level conditional posterior $P(g^L_{t+T_p} | x_{1:t+T_p}^{h,r}, \textrm{Pa}(g^L_{t+T_p}))$;

            \Else
                \State Perform downward posterior propagation $P(g^{l}_{t+T_p} | x_{1:t+T_p}^{h,r})$ from $l+1$-level (Equation~\ref{eq:hit_downward_posterior_propagation});
            \EndIf
        \EndFor 
        \State Return posteriors at all levels $P(g^{l}_{t+T_p} | x_{1:t+T_p}^{h,r}), l=1,2,\ldots, L$ to the downstream pipeline;
    \EndFor
\end{algorithmic} 
\end{algorithm*}

\section{Collaborative Assembly Use Case}
\label{sec:hit++}
We present an application of our HIT algorithm for a collaborative assembly task.

\subsection{Collaborative Assembly Scenario}\label{sec:architecture-robot}

Consider a scenario where a human and a robot work together on an assembly task in close proximity, assembling four pairs of Misumi Waterproof E-Model Crimp Wire Connectors~\cite{kimble2020benchmarking}. These connectors feature asymmetrical shapes that require precise alignment and insertion of female and male parts to achieve tight clearances. At the beginning of each experiment trial, the female parts are individually placed in four square regions corresponding to the designated task locations $\{0, 1, 2, 3\}$, as depicted in Figure~\ref{fig:abnormal}. The male parts are initially located in a rectangular preparation area (“\textit{prep}” in Figure~\ref{fig:abnormal}) closer to the human. In this collaborative setting, each collaborator focuses on tasks suited to their respective strengths: the human leverages dexterity and adaptability to pick up male parts from the preparation area and align them precisely with female parts, while the robot applies its greater force capability to execute the push action for part assembly. The human leads the task by determining the sequence of part alignment. The robot is required to infer this sequence based on its knowledge of predefined task-intention locations to effectively perform the subsequent pushing actions. By tracking the human's task-level intention, the robot can autonomously push parts in parallel with the human's alignment—a mode of collaboration we refer to as Coexistence Mode. 

The assembly process presents certain challenges. The robot may fail to push at an optimal position required for a tight clearance, or an aligned male part may become dislodged due to table vibrations, resulting in assembly failure. We assume that the robot is incapable of reliably identifying these failures, as robots today still lack sufficient capability to perform high-quality inspection tasks. Instead, the human collaborator needs to visually inspect the outcome and physically guide the robot to the failure locations for successful task completion. We refer to this collaborative recovery process as Cooperation Mode. 

\subsection{HIT System Overview}

\begin{figure*}[t]
    \centering
    \includegraphics[width=0.96\linewidth]{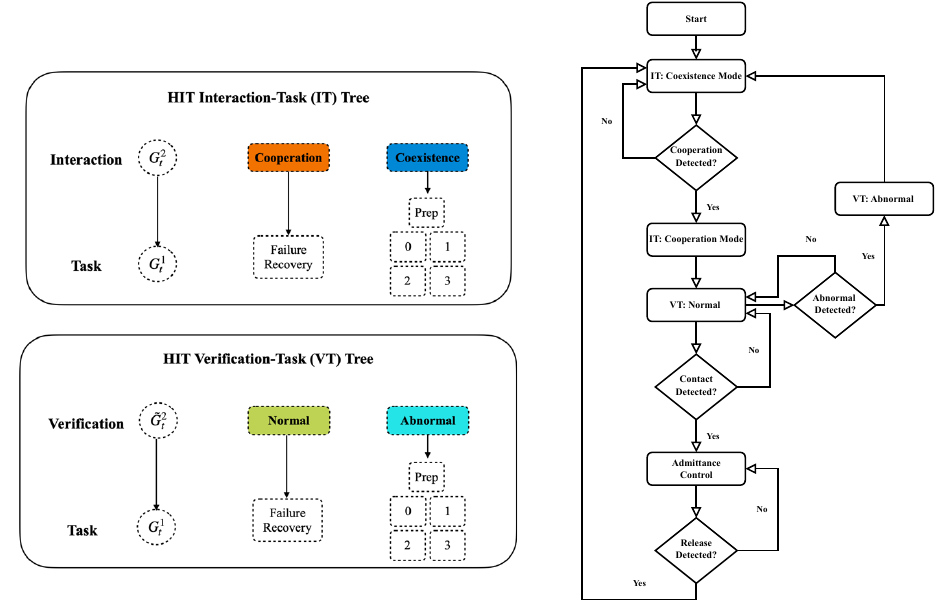}
    \caption{\textbf{Illustration of our HIT-ITVT trees (left) and the flow chart for adaptive mode switching (right).} Our HIT-ITVT system consists of two two-layered trees: HIT-IT (top left) and HIT-VT (bottom left). The system dynamically switches between the two intention trees based on the current execution context as shown in the flow chart. HIT-IT serves to identify human's intended task sequence (task intention $G^1$) and detect transition between Coexistence Mode and Cooperation Mode (interaction intention $G^2$). HIT-VT monitors to determine whether the inferred cooperation is correct (\textit{Normal}) or incorrect (\textit{Abnormal}) (verification intention $\tilde{G}_2$) until the human confirms cooperation through physical contact.}
    \label{fig:HIT-ITVT-structure}
    \vspace{-10pt}
\end{figure*}
Figure~\ref{fig:hit-system-overview} presents an overview of our HIT system designed for the collaborative assembly scenario. To enable tracking of multiple types of human intentions, the system integrates multimodal sensing, including human motion tracking through vision-based perception, force/torque sensing for detecting physical hand guidance, and the robot's own states. Our system models and tracks human intentions across three semantic levels, as detailed in Figure~\ref{fig:HIT-ITVT-structure}: task-level, interaction-level, and verification-level intention. The task-level intention refers to the human's intended reaching target. In Coexistence Mode, this target corresponds to one of the five predefined square task locations including preparation area (\textit{prep}). In Cooperation Mode, the target is a dynamic goal region centered around the robot's end-effector where the human reaches out to activate manual guidance for failure recovery. The full set of task-level intentions is defined as $ G^1=\{0, 1, 2, 3, \textit{prep}, \textit{failure recovery}\}$. The interaction-level intention corresponds to the interaction mode $G^2=\{\textit{Cooperation, Coexistence}\}$. The verification-level intention assesses whether the human's actual behavior aligns with the tracked \textit{Cooperation} intention, enabling the system to self-correct misclassifications through the verification intention state $\tilde{G}^2=\{\textit{Normal, Abnormal}\}$. Based on the inferred intentions, the robot dynamically switches between Coexistence and Cooperation Modes to ensure seamless and effective HRC.

\subsection{Hierarchical Intention Switching and Validation}\label{sec:architecture-switch}

\begin{figure*}[t]
\centering

\vspace{-0.6cm}
\begin{minipage}[t]{0.2\textwidth}
  \includegraphics[width=\linewidth]{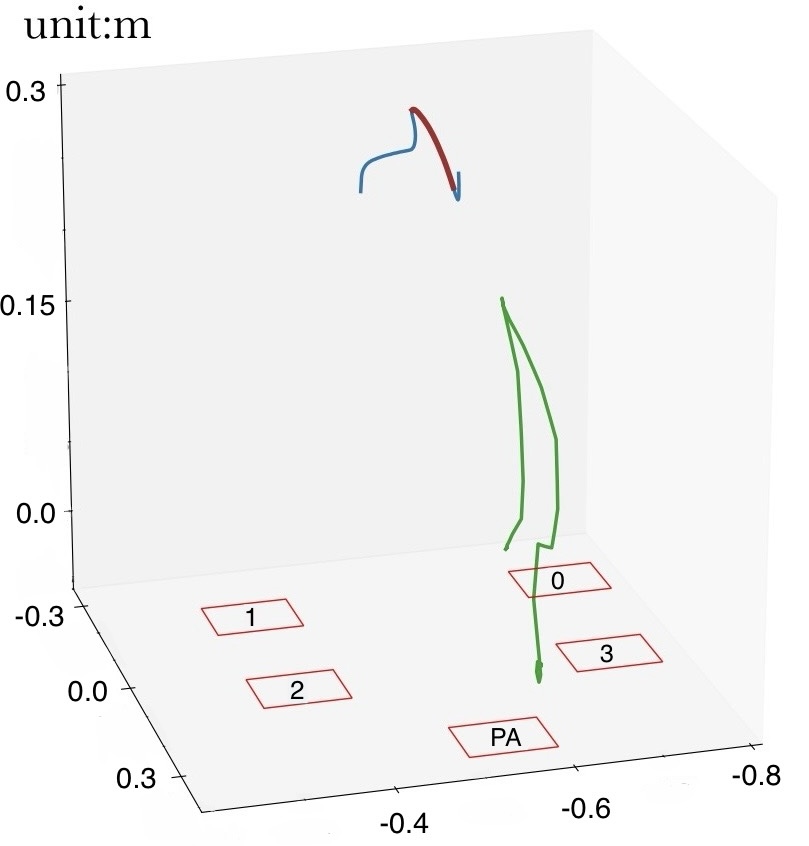}
\end{minipage}
\begin{minipage}[t]{0.79\textwidth}
  \includegraphics[width=\linewidth]{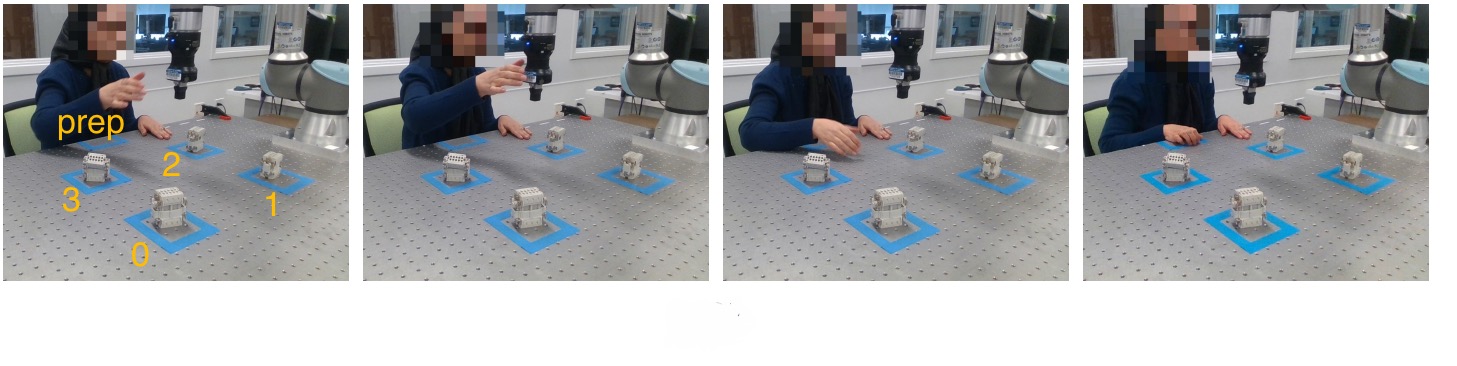}
\end{minipage}
\vspace{-0.2cm}
\centering (a)

% % ---------- Horizontal Dashed Line ----------
% \begin{tikzpicture}
%     \draw[dashed] (0,0) -- (\textwidth,0);
% \end{tikzpicture}

% ---------- ArXiv-safe dashed horizontal line ----------
\dashedline

\begin{minipage}[t]{0.20\textwidth}
  \includegraphics[width=\linewidth]{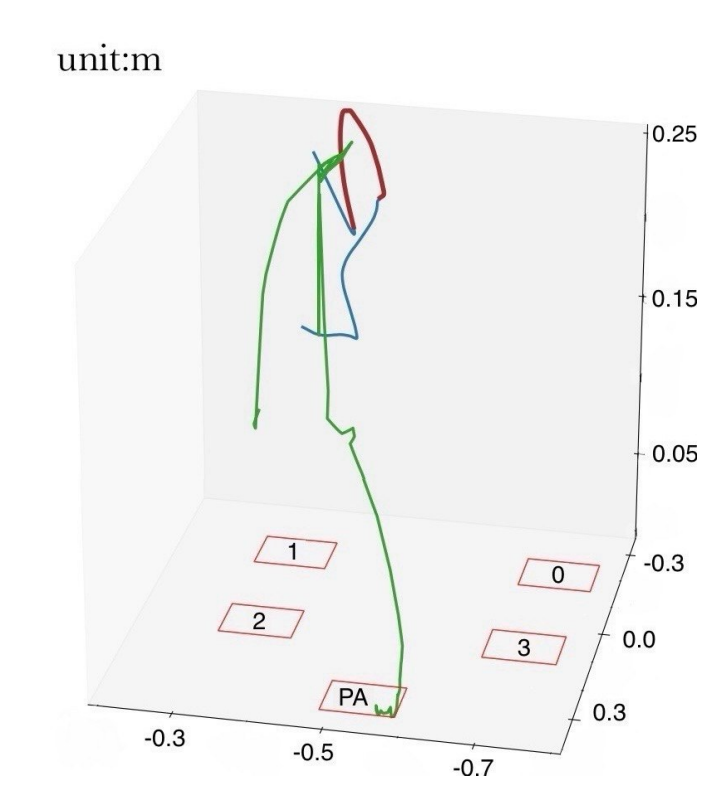}
\end{minipage}
\begin{minipage}[t]{0.79\textwidth}
  \includegraphics[width=\linewidth]{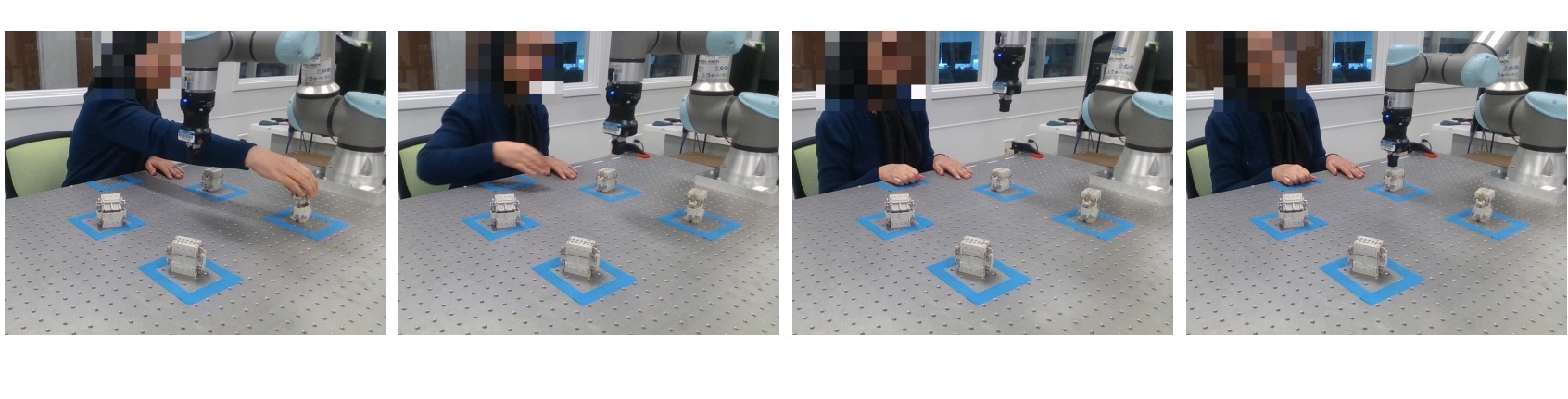}
\end{minipage}

\vspace{-0.2cm}
\centering (b)

\caption{\textbf{Examples of Self-Correcting HIT-ITVT System in Two Abnormal Cases.} Leftmost figures in \textbf{(a,b)}: Trajectories of human's hand (\textcolor{ForestGreen}{green}) and robot's end-effector (\textcolor{blue}{blue} for Coexistence Mode and \textcolor{red}{red} for abnormality detection).  Snapshot of (a): The human mistakenly believes a failure has occurred and reaches toward the robot. Interpreting this gesture as an intent to cooperate, the robot begins to approach the human hand. However, upon realizing there is no failure, the human withdraws their hand, and the robot, operating under HIT-ITVT system, returns to its default position. Snapshot of (b): As the human reaches toward the upper-left part to perform alignment, their hand passes near the robot's end-effector, which is incorrectly detected as a \textit{Cooperation} attempt. Detecting that the hand continues moving away, the HIT-ITVT system self-corrects, allowing the robot to return to its autonomous task—pushing the lower-left part.}
\label{fig:abnormal}
\end{figure*}

As illustrated in Figure~\ref{fig:HIT-ITVT-structure}, our HIT system in this use case employs two two-layered trees: the Intention-Task (IT) tree and the Verification-Task (VT) tree. The IT tree consists of a lower layer representing task-level intentions and a higher layer capturing interaction-level intentions. In contrast, the VT tree shares the same task-level intention layer but replaces the upper layer to track verification-level intentions, allowing the system to detect and correct potential misclassifications of interaction intentions.

The HIT-ITVT system dynamically alternates between IT and VT trees based on the current execution context and system states as illustrated in the flowchart in Figure~\ref{fig:HIT-ITVT-structure}. HIT-IT is active during Coexistence Mode (interaction intention in $G^2$) and primarily serves to identify the human's intended task sequence (task intention $G^1$) and detect transition to Cooperation Mode (interaction intention in $G^2$). Once Cooperation Mode is detected, the system switches to HIT-VT, and the robot shifts from maintaining a safe distance to actively approaching the human wrist, signaling its readiness for hand guidance. 

Occasionally, the robots might misinterpret the human's motion near the end-effector as an intent to cooperate. The false \textit{Cooperation} intentions may arise when the human passes near the robot's end-effector en route to another location or when the human temporarily reverts to realign the parts (\textit{Coexistence}) while moving towards the robot as illustrated in Figure~\ref{fig:abnormal}. To prevent unsafe or unnecessary interactions, HIT-VT monitors verification intention $\tilde{G}_2$ to determine whether the inferred \textit{Cooperation} is correct (\textit{Normal}) or incorrect (\textit{Abnormal}) until the human confirms \textit{Cooperation} through physical contact. If an abnormal intention is detected during the approach phase, the system immediately switches back to HIT-IT and resume Coexistence Mode.

Once physical contact is detected without \textit{Abnormal} detection, the robot transitions to admittance control to enable hand guidance. During this hand-guidance (Cooperation Mode), intention tracking is temporarily paused because the human collaborator has full control of the robot. Once the human releases the robot, HIT-IT is reactivated to resume autonomous intention tracking and task execution (Coexistence Mode).

\subsection{Intention-aware Human Behavior Model} \label{sec:behaviormodel}

To compute the bottom-level measurement likelihood $p(x_{t+1:t+T_p}^{h,r} | x_{1:t}^{h,r},g^{1}_{t+T_p})$ in Equation~\ref{eq:hit_single_layer_measurement_model}, we implement the probabilistic intention-aware human behavior model, where the human wrist position variable $\bm{X}_t$ is modeled as a Gaussian random variable. In the context of the collaborative assembly task, the goal position $G$ is the task-level goal-reaching intentions (i.e. reaching for a specific part), the low-level intentions for both IT and VT, and  is represented as a Gaussian. Since the model is designed to capture short-term human behavior over the intention update horizon $T_p$, we assume the goal $\bm{G}$ remains static during this period as discussed in Section~\ref{sec:hit}. With Gaussian process noise $\bm{W}_t$, we define the intention-aware human behavior model as follows:  

\begin{equation}\label{eq:intention_aware_human_behavior_model_1}
    \vspace{-0.2cm}
    \begin{aligned}
    \bm{X}_{t+1}^h = \bm{X}_t^h + & \alpha_t\left(\bm{G} - \bm{X}_t^h\right)  + \bm{W}_t \\
    \bm{X}_t^h &\sim \mathcal{N}(\bm{\mu}_{t}, \bm{\Sigma}_{t}) \\
    \bm{G} &\sim \mathcal{N}(\bm{\mu}_G, \bm{\Sigma}_G) \\
    \bm{W}_t &\sim \mathcal{N}(\bm{0}, \bm{\Sigma}_W) \\
    \alpha_t & := \frac{v_t\, dt}{||\bm{\mu}_G - \bm{\mu}_{t}||} \\
    \end{aligned}
\end{equation}
where $v_t$ is average wrist speed estimated over a moving window, and $dt$ is the time step. We can derive that $\bm{X}_{t+1}$ remains Gaussian as shown below:

\begin{equation}\label{eq:intention_aware_human_behavior_model_2}
    \vspace{-0.2cm}
    \begin{aligned}
    \bm{X}_{t+1}^h &\sim \mathcal{N}(\bm{\mu}_{t+1}, \bm{\Sigma}_{t+1}) \\
    \bm{\mu}_{t+1} & = (1-\alpha_t) \bm{\mu}_{t} + \alpha_t \bm{\mu}_G\\
    \bm{\Sigma}_{t+1} = &
    (1-\alpha_t)^2 \bm{\Sigma}_{t} + \alpha_t^2 \bm{\Sigma}_G + \bm{\Sigma}_W + \\
    &\alpha_t(1-\alpha_t)\left(\textrm{Cov}(\bm{X}_t^h, \bm{G}) + \textrm{Cov}(\bm{G}, \bm{X}_t^h)\right) \\
    \textrm{Cov}(\bm{X}_{t+1}^h, \bm{G}) & = (1-\alpha_t) \textrm{Cov}(\bm{X}_{t}^h, \bm{G}) + \alpha_t \bm{\Sigma}_G \\
    \end{aligned}
\end{equation}

In practice, at each time step $t$, we obtain a human wrist position measurement $x_t$ which is used to set the Gaussian mean $\bm{\mu}_t$ and the variance $\bm{\Sigma}_t$ from a Kalman-Filter based human wrist tracker. Given the defined parameters $\bm{\mu}_G$, $\bm{\Sigma}_G$, and $\bm{\Sigma}_W$, along with the assumption $\textrm{Cov}(\bm{X}_{t}^h, \bm{G}) = 0$, we can iteratively compute the predictive distribution of the wrist position $\bm{X}_{t+1:t+T_p}^h$ over the horizon $T_p$. Following Equation~\ref{eq:hit_measurement_model_in_practice}, we compute the bottom-level measurement likelihood as:

\begin{equation}\label{eq:hit_measurement_model_in_practice}
\begin{aligned}
    p(x_{t+1:t+T_p}^{h,r} | x_{1:t}^{h,r}, g_{t+T_p})\!\!&=\!\!\prod_{\tau=1}^{T_p} \!\!p(x_{t+\tau}^{h} | x_{1:t+\tau-1}^{h,r}, x_{t+\tau}^r, g_{t+\tau}) \\
    &\approx \!\!\prod_{\tau=1}^{T_p}\!\!p(\bm{X}_{t+\tau}^{h} =  x_{t+\tau}^{h})\\
\end{aligned}
\end{equation}

It is worth noting that this implementation in Equation~\ref{eq:hit_measurement_model_in_practice} represents a practical adaptation of the theoretical formulation in Equation~\ref{eq:hit_single_layer_measurement_model}. We observed that directly enforcing $\bm{\mu}_{t+\tau}$ to be $x_{t+\tau}$ led to unrealistically small prediction errors, which caused overly aggressive updates to the posterior with excessive sensitivity during intention tracking. To mitigate this issue, we modified the measurement model to rely on the likelihood of the observed human state without conditioning on exact predictive means. This modification stabilizes the posterior update and leads to more robust intention tracking in practice.

\section{Implementation}
\label{sec:implementation}

We present the implementation details of our collaborative assembly use case.

\subsection{Robot and Perception Setup}\label{sec:architecture-robot}
\begin{figure}[h]
    \centering
    \includegraphics[width=\linewidth]{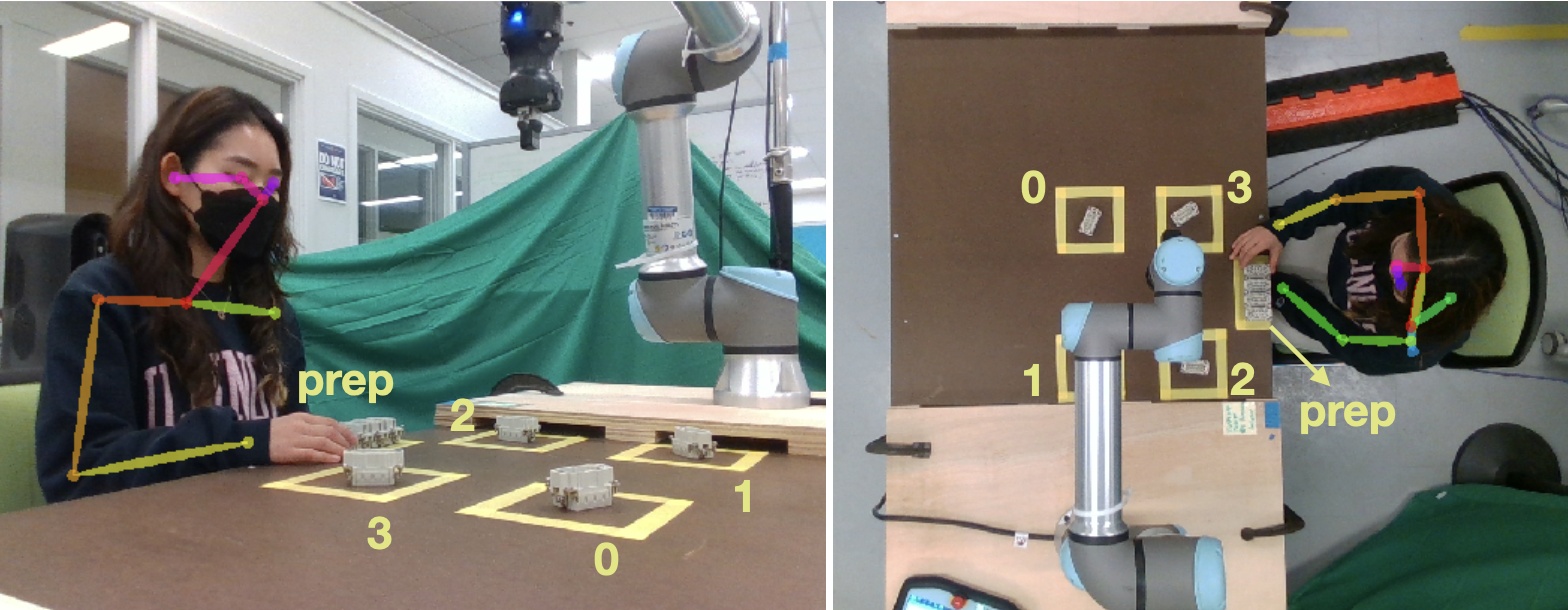}
    \caption{\textbf{Vision-Based Human Motion Tracking.} Human skeleton detection by OpenPose~\cite{cao2017realtime} on frames from a side-view (left) and a top-view (right) camera. Multiple cameras are used to address occlusion which happens frequently in close-proximity HRC.}
    \label{fig:human-tracking}
    \vspace{-15pt}
\end{figure}

We use the UR5e arm equipped with a Robotiq Hand-E Gripper. An embedded force and torque sensor measures the applied force at the end-effector at $500$~Hz. The robot is controlled by joint velocity commands and operates at a reduced speed ($30\%$ of its  maximum) to ensure safety. We also use two Intel RealSense RGBD cameras, which provide complementary top-down and side views of the shared space to alleviate occlusions as shown in Figure~\ref{fig:human-tracking}. Human skeletons are detected using OpenPose~\cite{cao2017realtime} from both camera views. We apply Kalman Filter to these detections and track 3D positions of human right wrist at $30$~Hz. All modules are executed using Robot Operating System.

\subsection{Dual-Mode Collaboration Strategy}\label{sec:architecture-planning}

\textbf{Coexistence Module} is developed for concurrent task execution without interruption by tracking the probability distribution over task-level intentions, focusing on the part alignment locations $g_t^1 \in \{0, 1,2,3, \textit{prep}\}$. If the probability of the most likely task intention remains above $0.8$ for at least $2$~\si{seconds}, the human collaborator is considered to be actively aligning a part in the corresponding task region. This alignment state is maintained until the probability drops below $0.25$, indicating completion of the alignment. Each newly recognized alignment is added to a queue of task intentions, and the robot plans trajectories to perform the pushing actions in a first-come, first-serve manner. To prevent redundant actions, each task intention is added to the queue only once. The robot executes the push only once per task intention region unless manual guidance from the human triggers a failure recovery. If the task intention probability exceeds $0.8$ for less than $2$~\si{seconds}, the event is regarded as a false positive or transient motion (e.g. the human passing by the region en route to another location), and the alignment is not registered. In this way, the $2$-\si{second} duration serves as a filtering threshold to reject spurious detections and ensure robustness. 

Given the close proximity between the robot and the human, a collision-avoidance mechanism is also integrated to ensure human safety and minimize interference with the human's movements. Specifically, the robot generates a collision-avoidance trajectory shaped by a repulsive potential field~\cite{khatib1986real}, steering the robot's end-effector away from the human wrist. Force control is also incorporated to reliably detect the completion of each pushing action. %\textcolor{red}{need to mention task planning.} 
To simulate potential failures typically encountered in more complex, real-world tasks, a small uniform random noise is added into the robot's pushing position. This controlled perturbation allows us to evaluate the robustness of the assembly process under imperfect execution.

\textbf{Cooperation Module} facilitates manual guidance by the human collaborator to address and recover from assembly failures. \textit{Cooperation} intention is initially detected by the IT tree ($g^1_t={\textit{failure recovery}}$) when the human hand approaches the robot's end-effector, and is subsequently verified by the VT tree. If the probability of \textit{Cooperation} intention surpasses $0.9$ for more than $0.5$~\si{second}, the system transitions to the Cooperation Module and activates VT to determine whether the detected cooperative intention is correct (\textit{Normal}) or incorrect (\textit{Abnormal}). If VT identifies the human's true intention as targeting one of the predefined part alignment regions or the preparation area ($\{0, 1,2,3, \textit{prep}\}$)—indicating an \textit{Abnormal} case—with probability above $0.8$ for more than $1$~\si{second}, the system reverts back to the IT tree and resumes with Coexistence Mode. 

Upon detection of \textit{Cooperation} by the IT tree, the robot switches from collision avoidance to approaching behavior, using an attractive potential field to guide the robot's end-effector toward the human wrist—clearly signaling readiness for hand guidance. Once physical contact is established (contact force$>20$~\si{N}), admittance control is activated, allowing the human to intuitively guide the robot to the desired ready-to-push position. When the human releases the robot at the target position and the measured force falls below $3$~\si{N} for more than $1.5$~\si{seconds}, the robot immediately executes a pushing action at that position to complete the recovery from assembly failure. To ensure a smooth transition between Coexistence and Cooperation Modes, a running mean of joint velocities is maintained, preventing discontinuities in control input~\cite{cacace2018shared}.

\subsection{HIT Layers}\label{sec:architecture-layers}
\textbf{Low-Level Tracking}
In both IT and VT, the low-level intentions correspond to task intentions associated with four assembly locations and preparation area: $\{0,1,2,3, \textit{prep}\}\in G^1$. We consider that these task intentions remain the same for $T_p=1/30$~\si{second} and track at a frequency of $30$~\si{Hz} using the Gaussian human behavior model described in Section~\ref{sec:behaviormodel}. The model takes as input the observed $3$D human wrist trajectory $x^h_t$ and predefined intention regions $G^1$, and outputs a probability distribution over possible task intentions. We set both the wrist position variance $\bm{\Sigma}_t$, and goal variance $\bm{\Sigma}_G$ to $0.05$~\si{\meter^2}. Additionally, we define a dynamic process noise covariance $\bm{\Sigma}_W$ based on the human wrist speed $v_t$:
\begin{equation}\label{eq:gaussian_noise}
\bm{\Sigma}_W = 
\begin{cases}
\Vert \mu_t-\mu_G \Vert^2, & \text{if } v_t < 0.2 \text{\si{m/s}} \\
0.5, & \text{otherwise}
\end{cases}
\end{equation}
A distance-based noise term is used when the wrist moves slower than $0.2$~\si{m/s} while constant noise term is applied otherwise. This distance-based term accounts for greater uncertainty in wrist position when the wrist is far from the goal, thereby improving robustness in intention tracking. In particular, this formulation mitigates the effect of a very small $\alpha_t$, which would otherwise diminish the influence of the goal directed term $\alpha_t\left(\bm{G} - \bm{X}_t^h\right)$ and introduce noise into the intention estimate when the wrist is moving slowly. To compute the prediction in Equation~\ref{eq:hit_prediction_conditioned_on_parent_intention}, we set the human's momentum probability $\kappa^1=0.99$ in Equation~\ref{eq:hit_transition_model_conditioned_on_parent_intention}.

\textbf{High-Level Intention Tracking}
The high-level intention corresponds to either interactive intentions, $\{$\textit{Coexistence},  \textit{Cooperation}$\}$, tracked by the IT tree, or verification intentions,  $\{\textit{Normal} , \textit{Abnormal}\}$, tracked by the VT tree. Similar to low-level tracking, we consider that these high-level intentions remain the same for $T_p=1/30$~\si{second} and track at a frequency of $30$~\si{Hz}. The high-level intentions are also inferred using the same Gaussian human behavior model as in low-level tracking, with identical parameter settings. As presented in Figure~\ref{fig:HIT-ITVT-structure}, tracking for \textit{Coexistence} and \textit{Abnormal} relies on fixed goal regions ($\{0,1,2,3, prep\}\in G^1$), and tracking for \textit{Cooperation} and \textit{Normal} uses a dynamic goal region centered around the robot's end-effector ($\{\textit{failure recovery}\}\in G^1$). This hierarchical structure facilitates bidirectional information flow as described in section~\ref{subsec:hit}. When the inferred low-level intention $g^1\in G^1$ corresponds to the robot's end-effector area for longer than the stay-time threshold (in Section~\ref{sec:architecture-planning}), the IT tree detects \textit{Cooperation} and activates the VT tree for verification. If the human continues moving toward the end-effector, the intention is classified as \textit{Normal}. If movement shifts toward fixed task regions ($\{0,1,2,3, prep\}$), it is classified as \textit{Abnormal}, and the system switches to IT tree with Coexistence Mode. Similar to low-level tracking, $\kappa$ for both high-level intentions is set to $0.99$ when computing the transition model in Equation~\ref{eq:hit_transition_model_conditioned_on_parent_intention}.

\section{User Study}
\label{sec:userstudy}

\begin{figure*}
    \centering
    \includegraphics[width=\linewidth]{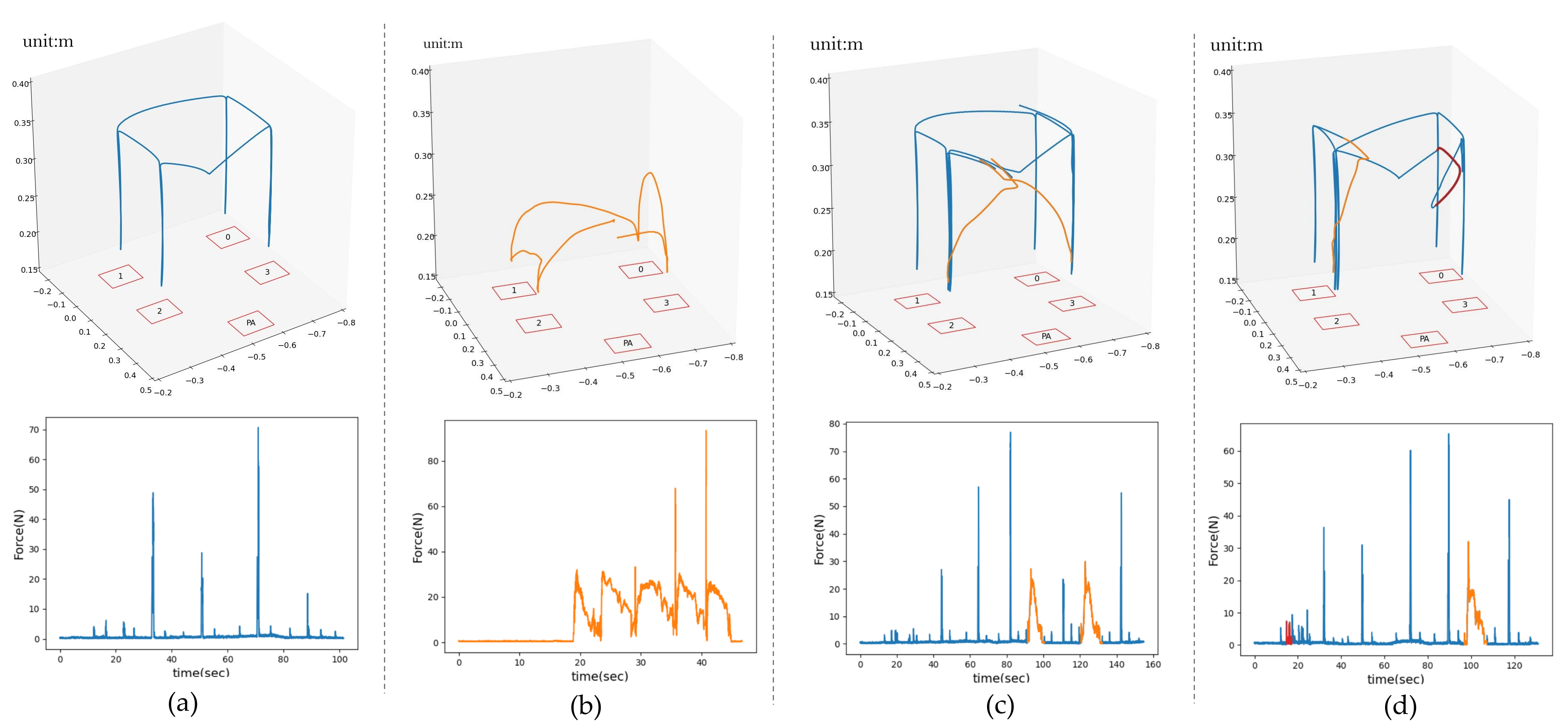}

    % \begin{tabular}{@{} c  c c c @{} }
    %     \includegraphics[width=0.23\linewidth]{Figures/coe/6_coe_traj_bold.png} &
    %      \includegraphics[width=0.23\linewidth]{Figures/col/6_col_traj_bold.png} &
    %     \includegraphics[width=0.24\linewidth]{Figures/ti/3_ti_traj_bold.png} &
    %     \includegraphics[width=0.22\linewidth]{Figures/tsti/8_tsti_traj_bold.png} \\
    %     \includegraphics[width=0.24\linewidth]{Figures/coe/6_coe_force_bold.png} &
    %     \includegraphics[width=0.24\linewidth]{Figures/col/6_col_force_bold.png} &
    %     \includegraphics[width=0.24\linewidth]{Figures/ti/3_ti_force_bold.png} &
    %     \includegraphics[width=0.24\linewidth]{Figures/tsti/8_tsti_force_bold.png} \\
    %     (a) &(b) & (c) &(d)
    % \end{tabular}
    \caption{\textbf{Trajectory and force visualization.} Robot trajectories (top) and applied forces (bottom) are shown for each module: (a) Coexistence Mode Baseline, (b) Cooperation Mode Baseline, (c) HIT-IT, and (d) HIT-ITVT. In all plots, trajectories and forces corresponding to Coexistence Mode are depicted in \textcolor{blue}{blue}, while those for Cooperation Mode are shown in \textcolor{orange}{orange}. In (d), \textcolor{red}{red} segments highlight instances of abnormal interaction—specifically, false \textit{Cooperation} detections. These segments exhibit shorter trajectories and lower force magnitudes compared to Cooperation Mode, reflecting the HIT-ITVT’s ability to detect anomalies and revert the robot to Coexistence Mode.}
    \label{fig:traj-force-plot}
\end{figure*}

We conducted a user study to evaluate our proposed HIT-based collaborative robot system on the collaborative assembly task introduced in Section~\ref{sec:hit++}. 

\subsection{Baselines and Measures}
We assessed the performance of our HIT-ITVT system against three baselines for HRC: Coexistence Mode Baseline, Cooperation Mode Baseline, and HIT-IT. In the Coexistence Mode Baseline, the robot executes low-level intention tracking, allowing autonomous concurrent task execution alongside the human but without the capability for failure recovery. The Cooperation Mode Baseline relies solely on admittance control, making the robot passively compliant; continuous human guidance is required to maneuver the robot and perform pushing actions. These two uni-modal systems are chosen as baselines because they are prevalent solutions in industry~\cite{villani2018survey, michalos2018seamless}. Additionally, to specifically evaluate the effect of the self-correcting feature provided by HIT-VT, we also compare our HIT-ITVT against HIT-IT. 
Evaluation metrics include (1) completion time,
(2) robot path length, (3) human path length, (4) average force exerted by the human, and (5) total human energy expenditure. We also report the number of assembly failures and the number of \textit{Abnormal} occurrences as these factors influence the performance metrics. Note, however, that HIT-IT and HIT-ITVT share the same condition for assembly failures and \textit{Abnormal} occurrence counts, as both use the same IT trees that trigger these events. The number of assembly failures and instances of \textit{Abnormal} were manually recorded.

\subsection{Participants}
We recruited $10$ participants ($5$ females; $5$ males; ages $29.8 \pm 9.4$ years) for the user study. All participants provided informed written consent prior to the experiment (IRB $\# 24-0180$). At the beginning of the study, participants completed a presurvey about their demographics and perceptions of the robots as a collaborator. We employed Wilcoxon signed-rank tests to evaluate the participants' attitudes towards collaborative robots. Results from the presurvey indicated that although participants generally trust the robot as a collaborator for simple tasks, their willingness to engage in collaboration was significantly lower than their trust ratings ($W=0.0, p<0.01$). This ambivalence seems to arise primarily due to their concern about the robot's limited capability to accurately interpret human intentions as indicated by a significant discrepancy between trust scores and participants' expectations regarding the robot's performance in intention recognition ($W=2.5, p<0.05$). 

\subsection{Procedure}
Before the experiment begins, participants were given sufficient time to familiarize themselves with the robot's key features, including collision avoidance (via a repulsive potential field), cooperative initiation (approaching the human using an attractive potential field), and manual guidance (through admittance control). Participants then interacted with the four modules—Coexistence Mode Baseline, Cooperation Mode Baseline, HIT-IT, and HIT-ITVT—six times each. We excluded trials if participants deviated from prescribed procedures, such as engaging in unrelated manual tasks, that could impact performance. After removing these invalid trials, we obtained the final dataset comprising $60$ trials for Coexistence Mode Baseline, $60$ for Cooperation Mode Baseline, $59$ for HIT-IT, and $57$ for HIT-ITVT. To reduce potential bias stemming from similarities between HIT-IT and HIT-ITVT (both featuring coexistence and cooperation functionalities, differing primarily in the verification module), we randomized the interaction order: five participants interacted with HIT-IT first, followed by HIT-ITVT, while the other five followed the reverse order. After completing interactions with each module set, participants filled out post-interaction surveys related to their experience and responded to brief interviews. 

\subsection{Results}
\begin{table*}[ht]
\caption{\textbf{Results of the ablative study.} We compare the performance of our HIT-ITVT system against three baselines for HRC: Coexistence Mode Baseline, Cooperation Mode Baseline, and HIT-IT. The mean and standard error of mean is presented for each metric category. The best (second best)
performance is denoted in bold
(underlined). We also report the number of assembly failures and the number of \textit{Abnormal} occurrences as these factors influence the performance metrics. Note, however, that HIT-IT and HIT-ITVT share the same condition for assembly failures and \textit{Abnormal} occurrence counts, as both use the same IT trees that trigger these events. The number (\#) of assembly failures and instances of abnormal behavior were manually recorded. `R.' and `H.' denote robot and human, respectively. 
}\label{table:ablative}
\begin{center}
\setlength\tabcolsep{3pt}
\begin{tabular}{lccccccccc}
    \toprule
    System & Completion Time (sec) & R. Path (m) & H. Path (m) & H. Force (N) & H. Energy (J) & Failures \# & Failure Recovery \# & Abnormal \#\\
    \midrule
    % \midrule
    Coexistence & $\underline{103.08\pm8.50}$ & $\underline{2.96\pm0.22}$ & $\mathbf{6.72\pm1.86}$ & N/A & N/A & $1.84$ &N/A & N/A\\
    Cooperation & $\mathbf{64.47\pm11.73}$ & $\mathbf{1.79 \pm 0.19}$ & $\underline{8.90\pm3.23}$ & $9.42\pm1.44$ & $31.83\pm7.53$ & $\mathbf{0.0}$ & N/A & N/A\\
    HIT-IT & $171.12\pm45.41$ & $5.54\pm1.82$ & $12.50\pm4.88$ & $\underline{1.33\pm0.69}$ & $\underline{13.39\pm10.10}$ & $\mathbf{0.0}$ & $\mathbf{1.61\pm0.12}$& $\mathbf{0.58\pm0.11}$\\
    HIT-ITVT & $167.14\pm33.52$ & $5.15\pm1.13$ & $12.55\pm4.36$ & $\mathbf{1.03\pm0.59}$ & $\mathbf{8.88\pm7.21}$ & $\mathbf{0.0}$ & $\underline{1.74\pm0.13}$ & $\underline{0.86\pm0.12}$\\
    \bottomrule
\end{tabular}
\end{center}
\vspace{-10pt}
\end{table*}

\begin{table*}[hb]
\caption{\textbf{Performance comparison between HIT-IT and HIT-ITVT for trials involving abnormal cases.} The mean and standard error of mean is presented for each metric category. The evaluation is based on trials that include at least one abnormal case: $21$ trials for HIT-IT and $33$ for HIT-ITVT. The best 
performance in each category is indicated in bold. We also report the number of assembly failures and the number of \textit{Abnormal} occurrences as these factors influence the performance metrics. depend on them. Note, however, that HIT-IT and HIT-ITVT share the same condition for assembly failures and \textit{Abnormal} occurrence counts, as both use the same IT trees that trigger these events. The number of assembly failures and instances of abnormal behavior were manually recorded.  %\textcolor{red}{Need to figure out how to match the text format of table to figure.}}
}
\label{table:abnormaldata}
\begin{center}
\setlength\tabcolsep{3pt}
\begin{tabular}{lccccccc}
    \toprule
    System & Completion Time (sec) & Robot Path (m) & Human Path (m) & Human Force (N) & Human Energy (J) & Failure Recovery \# & Abnormal \# \\
    \midrule
    % \midrule
    HIT-IT & $202.28\pm38.63$ & $6.71\pm1.41$ & $15.05\pm4.86$ & $1.72\pm0.65$ & $20.49\pm9.54$ & $\mathbf{1.62\pm0.20}$ & $1.62\pm0.13$ \\
    HIT-ITVT & $\mathbf{178.17\pm29.15}$ & $\mathbf{5.57\pm0.98}$ & $\mathbf{13.68\pm3.94}$ & $\mathbf{1.02\pm0.57}$ & $\mathbf{9.25\pm7.43}$ & $1.76\pm0.16$ & $\mathbf{1.48\pm0.11}$ \\
    \bottomrule
\end{tabular}
\end{center}
\vspace{-10pt}
\end{table*}
The quantitative evaluation results are summarized in Table~\ref{table:ablative}. The Coexistence Mode Baseline exhibits limited efficiency compared to Cooperation Mode Baseline because the robot must operate at lower speeds due to its close proximity to human
collaborator. Additionally, since the robot cannot independently inspect assembly failures, the Coexistence Mode Baseline offers only a single attempt to push each part. This constraint resulted in a relatively high average failure rate of $1.84$ per trial, whereas no assembly failures occurred in the other modules. Conversely, the Cooperation Mode Baseline achieves the shortest completion times at the cost of substantial human effort in terms of applied force and energy. As demonstrated in Figure~\ref{fig:traj-force-plot} (b), continuous human guidance is necessary to position the robot correctly and execute pushing actions. This constant manual guidance not only demands significant human energy but also increases fatigue risk. Moreover, abrupt forces during pushing demonstrate physical strain, potentially leading to workplace injuries.

In contrast, the HIT systems, HIT-IT and HIT-ITVT, effectively integrate coexistence and cooperation functionalities, leveraging the strengths of each mode. Both HIT systems completed tasks without any assembly failures and significantly reduced human applied force and energy compared to the Cooperation Mode Baseline, as human guidance is required only during failure recovery. Figure~\ref{fig:traj-force-plot} (c) highlights that Cooperation Mode is activated briefly only when needed, while most goal-reaching movements and all pushing tasks are performed autonomously by the Coexistence Module.

HIT-ITVT further improves upon HIT-IT by incorporating self-correcting intention prediction. Despite involving more frequent failure recovery events that require manual guidance from the human, HIT-ITVT results in less human-applied force and energy. Such benefits are particularly evident in trials involving abnormal cases, as summarized in Table~\ref{table:abnormaldata}. This improvement arises from the system's ability to correct misdetected \textit{Cooperation} intentions, relieving humans from the need to actively counteract incorrect robot approaches. Two representative \textit{Abnormal} scenarios in Figure~\ref{fig:abnormal} illustrates how HIT-ITVT responds to misclassified \textit{Cooperation} intentions. Despite a higher number of failure recovery events, HIT-ITVT demonstrates greater efficiency regarding completion time and reduces human workload, including  movements, applied force, and overall energy expenditure.

\subsection{Subjective Analysis}
\begin{table*}[ht]
\caption{\textbf{Results of post survey.} The participants evaluated their collaboration experience across multiple dimensions using $5$-point Likert scale. The mean and standard error of mean is presented for each metric category. The best (second best)
performance is denoted in bold
(underlined).}\label{table:survey}
\begin{center}
\setlength\tabcolsep{3pt}
\begin{tabular}{lccccccccc}
    \toprule
    & Satisfaction$\uparrow$ & Adaptability$\uparrow$ & Mental Fatigue$\downarrow$ & Physical Fatigue$\downarrow$ & Safety$\uparrow$ & Trust$\uparrow$ & Efficiency $\uparrow$ & Task Intent$\uparrow$ & Interact Intent$\uparrow$
    \\
    \midrule
    % \midrule
    Coexistence & $2.4 \pm 0.4$ & $2.2 \pm 0.3$ & $\underline{2.6 \pm 0.4}$ & $\mathbf{1.8 \pm 0.3}$ & $\underline{4.1 \pm 0.3}$ & $2.6 \pm 0.4$ & $1.6 \pm 0.3$ & $3.9 \pm 0.4$ & N/A\\
    Cooperation & $3.0 \pm 0.4$ & $1.5 \pm 0.3$ & $2.9 \pm 0.5$ & $3.7 \pm 0.4$ & $\mathbf{4.3 \pm 0.4}$ & $3.4 \pm 0.4$ & $2.0 \pm 0.3$ &  N/A &  N/A \\
    HIT-IT & $\underline{3.4 \pm 0.4}$ & $\underline{3 \pm 0.4}$ & $2.9 \pm 0.38$ & $2.8 \pm 0.3$ & $3.4 \pm 0.4$ & $\underline{3.5 \pm 0.3}$ & $\mathbf{2.9 \pm 0.4}$ & $\underline{4.1 \pm 0.3}$ & $\underline{3.7 \pm 0.3}$ \\
    HIT-ITVT & $\mathbf{3.9 \pm 0.3}$ & $\mathbf{3.5 \pm 0.3}$ & $\mathbf{2.2 \pm 0.3}$ & $\underline{1.9 \pm 0.3}$ & $3.7 \pm 0.3$ & $\mathbf{4.0 \pm 0.3}$ & $\underline{2.7 \pm 0.4}$ & $\mathbf{4.4 \pm 0.2}$ & $\mathbf{4.7 \pm 0.2}$ \\
    \bottomrule
\end{tabular}
\end{center}
\vspace{-10pt}
\end{table*}
Participants evaluated their collaboration experience across multiple dimensions using $5$-point Likert scale, including satisfaction, adaptability, fatigue (mental and physical), safety, trust, efficiency, and intention-related measures (task intention and interaction intention) as shown in Table~\ref{table:survey}. We employed linear mixed models to examine statistically significant differences among the four modules. 

The Cooperation Mode Baseline and Coexistence Mode Baseline showed strength in specific areas but exhibited major weaknesses in others. The Cooperation Mode Baseline achieved the highest perceived safety, notaly in comparison to HIT-IT ($se\!=\!0.425, t\!=\!-2.118, p\!<\!0.05$) but also resulted in elevated physical fatigue ($se\!=\!0.320, t\!=\!-2.815, p\!<\!0.01$). This elevated safety perception stems from participants’ full control over the robot in the Cooperation Mode Baseline, as noted by one of the participants: \textit{“I felt relatively safe because the robot wasn't moving on its own.”}. The Coexistence Mode Baseline also achieved higher safety score compared to the HIT algorithms, but these differences were not statistically significant ($p\!\!>\!\!0.05$ for HIT-IT and $p\!\!>\!\!0.1$ for HIT-ITVT). This outcome may be attributed to the increased behavioral complexity of the HIT models, which integrate both \textit{Coexistence} and \textit{Cooperation} strategies. Such complexity may reduce users’ perceived safety, as illustrated by one participant’s comment: \textit{“So compared to the previous two (Coexistence Mode and Cooperation Mode Baselines), I felt unsafe. I think because it (robot) was trying to both do collision avoidance and then, trying to cooperate in case I want to do cooperation.”}

The HIT-IT model demonstrated strong performance %, particularly in satisfaction, adaptability, and intention recognition, 
following HIT-ITVT. HIT-IT showed significantly higher efficiency ($se\!=\!0.328, t\!=\!-2.746, p\!<\!0.01; se\!=\!0.328, t\!=\!-3.967, p\!<\!0.001$) and adaptability ($se\!=\!0.336, t\!=\!-4.470, p\!<\!0.001;se\!=\!0.336, t\!=\!-2.384, p<0.05$) than both the Cooperation and Coexistence Mode Baselines, respectively. Moreover, HIT-IT highlighted less physical stress compared to the Cooperation Mode Baseline ($se\!=\!0.320, t\!=\!2.815, p\!<\!0.01$) and higher trust than the Coexistence Mode Baseline ($se\!=\!0.450, t\!=\!-2.000, p\!<\!0.05$). HIT-IT's failure recovery capability played the key role in such trust, as one of the participant noted that \textit{“In comparison to other model (Coexistence Mode Baseline), the good thing about this model is that there was an option for me to intervene and fix the failure.”}. These results suggest that HIT-IT successfully combines the strengths of the Cooperation and Co-existence Mode Baselines and complements the respective limitations. 

\begin{figure*}[th!]
    \centering
    \includegraphics[width=\linewidth]{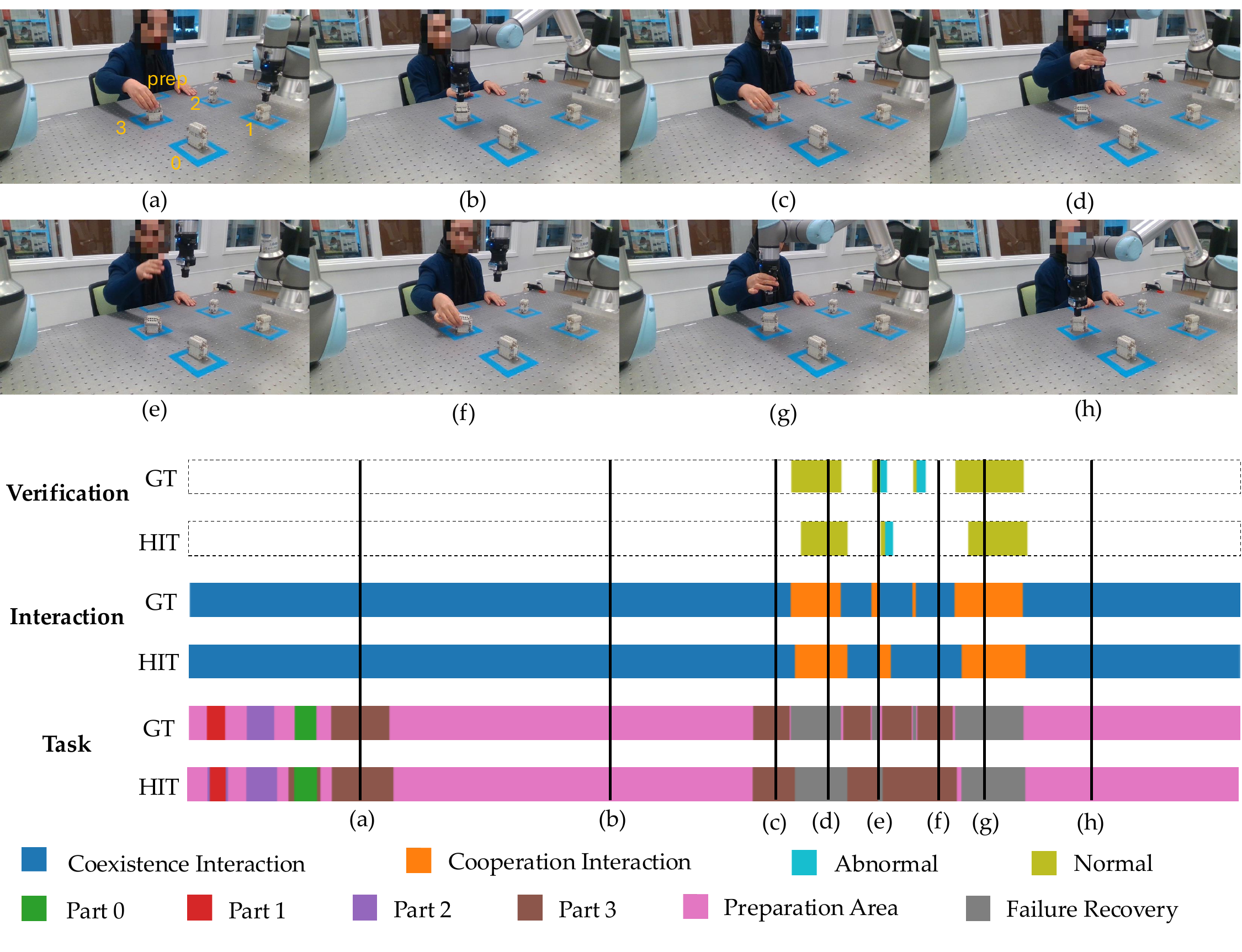}
    \caption{\textbf{Illustration of a participant interacting with the HIT-ITVT model.} The intention tracking bars display the most likely inferred intentions from our HIT-ITVT (detnoted as HIT) alongside the corresponding ground truth (denoted as GT) at each time step. The accompanying snapshots depict a sequence highlighting the model's failure recovery and self-correcting capabilities. (a) The human aligns part $3$. (b) The robot fails to assemble the part. (c) The human realigns the part. (d) The human hand-guides the robot to the realigned part, but the part becomes dislodged due to table vibration. (e) The human reaches out to the robot to cooperate (\textit{Normal} intention followed by \textit{Cooperation} intention), but the part is dislodged again. (f) Before making contact with the robot, the human returns to the part location (\textit{Abnormal} intention) to correct the alignment once more. (g) After correcting the alignment, the human hand-guides the robot to the appropriate ready-to-push position. (h) The robot recovers the failure by pushing the part again. For visualization purposes, the stay-time threshold, used in the HIT-ITVT system to filter out transient misclassifications, is ignored in this diagram. As a result, temporary detections are included in this plot, such as a brief segments of transitive task intentions (Part 2, 3, and Preparation Area). Also, for simplicity, periods in which the human moves between task locations are labeled as ``Preparation Area" although the underlying intention may vary. In this trial, the model achieved frame-wise accuracies of $90.0 \%$ for task intention, $87.8\%$ for interaction intention, and $95.5\%$ for verification intention.} 

    \label{fig:HIT_ITVT_ribbon}
    \vspace{-10pt}
\end{figure*}

Among all models, HIT-ITVT demonstrated the most robust overall performance. HIT-ITVT received significantly higher ratings in satisfaction ($se\!=\!0.461, t\!=\!-1.953, p=0.051; se\!=\!0.461, t\!=\!-3.255, p\!<\!0.01$), adaptability ($se\!=\!0.336, t=-5.960, p\!<\!0.001; se=0.336, t\!=\!-3.874, p\!<\!0.001$) compared to the Cooperation and Coexistence Mode Baselines, respectively. The failure recovery capability of HIT-ITVT fostered building higher trust in robot ($se\!=\!0.450, t\!=\!-3.111, p\!<\!0.01$) compared to the Coexistence Mode Baseline, while its self-correcting mechanism contributed to less physical stress ($se=0.320, t=2.815, p<0.01$) and higher interaction intent score ($se=0.501, t=-1.996, p<0.05$) relative to HIT-IT. One participant reflected on the contrast between HIT-ITVT and HIT-IT, stating, \textit{“It was the unsafe situation that I was not intending to cooperate at that specific time, but I think it was like misdetecting that (cooperative intention) and that (robot) was like coming to me.”} This feedback highlights the importance of HIT-ITVT’s self-correction mechanism, which enables to recover from misinterpreted intentions and improves the human’s perception of the robot’s ability to recognize cooperative intent. A demonstration of a participant interacting with the HIT-ITVT model is illustrated in Figure~\ref{fig:HIT_ITVT_ribbon}.

\subsection{Discussion}

Our user study results demonstrate that the proposed HIT systems, HIT-IT and HIT-ITVT, effectively integrate the benefits of Coexistence and Cooperation Mode. Unlike the Coexistence Mode Baseline, which operates conservatively due to proximity constraints and lacks failure recovery capabilities, the HIT systems support both autonomous task execution (e.g. part insertion) and occasional hand guidance for failure recovery. These features reduce failure rates and lessens human workload, including applied force and energy expenditure. Compared to the Cooperation Mode Baseline, which achieves faster task completion but demands continuous human guidance leading to higher physical fatigue, the HIT systems offer a more collaboration-friendly alternative by minimizing unnecessary human intervention. Notably, HIT-ITVT further improves interaction quality and efficiency through a self-correcting intention tracker that enables timely recovery from misdetections and maintains alignment with evolving human intent.

Subjective evaluations further highlight the advantages of HIT-ITVT in enhancing user experience. HIT-ITVT consistently outperformed other models in satisfaction, adaptability, trust, and intention recognition, while significantly reducing perceived fatigue. Participants especially appreciated the robot’s responsiveness and ability to recover from mistakes—capabilities that distinguish HIT-ITVT from both HIT-IT and baseline systems. One of the participants stated, \textit{“I can just do things normally (without any interruptions). There wasn't any case of a failed cooperation detection that I had to resolve or had to do anything more for. So it's much mentally less taxing than the previous experiment (HIT-IT)”}. Although the subjective safety scores were slightly lower for HIT systems due to their higher behavioral complexity, participants expressed greater trust in their autonomy and collaborative intelligence. These findings suggest that modeling and tracking hierarchical human intentions along with mechanisms for self-correction and failure recovery play a critical role in promoting effective, comfortable, and resilient HRC.

\section{Conclusions and Future Directions}
\label{sec:conclusion}

In this paper, we present the HIT algorithm to support safe, efficient, and adaptive HRC. HIT models human behavior with a dynamic hierarchy of human intentions and employs vertically stacked Bayesian filters to continuously track these intentions in real time. In a collaborative assembly use case, the HIT-based collaborative robot system is instantiated as two switching intention trees, Interaction-Task and Verification-Task intention trees (HIT-ITVT), which allows the robot to reason over human intentions at multiple levels and seamlessly transition between Coexistence and Cooperation Modes. By capturing human intentions across levels of task, interaction, and verification, the HIT-ITVT system allows the robot to align actions with human goals in complex and evolving collaborative scenarios. Our experiments and user studies show with this multi-level understanding, the robot can adapt to changing human behavior, recover from misdetections, and build user trust through smooth and reliable interaction and robust performance.

The proposed HIT algorithm open several promising avenues for future research in HRC. First, the current formulation of HIT assumes a fixed, pre-determined set of potential intentions in trees. Extending HIT to support an intention tree with dynamic structures, where new intentions can be added and old ones removed, would be critical for enabling the robot to help the human user in life-long, open-ended scenarios without having to explicitly split the collaboration into a pre-defined sequence of independent tasks. Second, HIT is generic and flexible to be adapted to arbitrary HRC tasks. It would be interesting to augment components of HIT such as transition matrices and measurement models with large language models and vision language models~\cite{huang2024lit}. This integration has the potential to enable robot agents to interpret and respond to human intentions in unstructured, open-world settings. In such settings, intentions cannot be pre-defined, but defined by reasoning from behavior context and the surrounding environment at run time. Third, the present work is focused on dyadic collaboration between a single robot and a single human. An extension of HIT can be a generalization to multi-human-multi-robot scenarios, which would involve research on heterogeneous agents, modeling interaction graphs, and optimizing task allocation and coordination within teams of collaborative robots and human partners.
\section{Acknowledgement}
\label{sec:acknowledge}
This work was supported by the National Science Foundation under Grant No. 2143435.


% \section{References Section}
\bibliographystyle{IEEEtran}
\bibliography{bib}

\vfill

\end{document}